\documentclass[twoside,11pt]{article}

\usepackage{blindtext}

\usepackage{jmlr2e}

\usepackage{lastpage}

\usepackage{times}  
\usepackage{helvet}  
\usepackage{courier}  
\usepackage{graphicx} 
\usepackage{natbib}  
\usepackage{caption} 
\usepackage{algorithm}
\usepackage{algorithmicx}
\usepackage[noend]{algpseudocode}

\usepackage{newfloat}
\usepackage{listings}
\usepackage{hyperref}

\usepackage{amsmath,amssymb,amsfonts}
\usepackage{graphicx}
\usepackage{textcomp}
\usepackage{xcolor}
\def\BibTeX{{\rm B\kern-.05em{\sc i\kern-.025em b}\kern-.08em
    T\kern-.1667em\lower.7ex\hbox{E}\kern-.125emX}}

\usepackage[utf8]{inputenc} 
\usepackage[T1]{fontenc}    
\usepackage{url}            
\usepackage{booktabs}       
\usepackage{makecell}       
\usepackage{amsfonts}       
\usepackage{nicefrac}       
\usepackage{microtype}      
\usepackage[dvipsnames]{xcolor}

\usepackage{mathtools}
\usepackage{thmtools}
\usepackage{thm-restate}
\usepackage{mathbbol}

\usepackage[]{caption}
\usepackage{subcaption}
\usepackage{float}
\usepackage{graphicx,xcolor,graphics,url}
\usepackage{verbatim,ifthen,alltt,array}
\usepackage{latexsym,dsfont}
\usepackage{booktabs,multirow}
\usepackage{enumitem}
\usepackage{xparse}

\usepackage{xspace}

\usepackage{xfrac}
\usepackage[framemethod=TikZ]{mdframed}
\usepackage{nicefrac}       

\usepackage{stmaryrd}
\usepackage{textcomp}
\usepackage{siunitx}
\usepackage{xpatch}
\usepackage{comment}
\usepackage{relsize}

\usepackage[capitalize,noabbrev,nameinlink]{cleveref}

\usepackage{pgf}
\usepackage{tikz}
\usepackage{forest}
\usepackage[implicantcolors={red,red,red,red,green,green,green,green},label=corner]{karnaugh-map} 

\usetikzlibrary{arrows,decorations.pathmorphing,decorations.pathreplacing,decorations.footprints,fadings,calc,trees,mindmap,shadows,decorations.text,patterns,positioning,shapes,matrix,fit}
\usetikzlibrary{arrows,shadows,backgrounds}
\usetikzlibrary{arrows.meta}
\usetikzlibrary{positioning,fit}
\usetikzlibrary{automata}
\usetikzlibrary{matrix}
\usetikzlibrary{shapes.symbols,shapes.misc,shapes.arrows}
\usetikzlibrary{matrix,chains,scopes,decorations.pathmorphing}
\usetikzlibrary{bayesnet}
\usetikzlibrary{tikzmark}

\jmlrheading{23}{2025}{1-\pageref{LastPage}}{9/25; Revised ?/??}{?/??}{21-0000}{J. Marques-Silva and A. Ignatiev}

\firstpageno{1}

\usetikzlibrary{circuits.logic.US}

\usetikzlibrary{arrows}
\usetikzlibrary{automata}
\usetikzlibrary{backgrounds}
\usetikzlibrary{calc}
\usetikzlibrary{chains}
\usetikzlibrary{circuits.logic.US}
\usetikzlibrary{decorations.footprints}
\usetikzlibrary{decorations.markings}
\usetikzlibrary{decorations.pathmorphing}
\usetikzlibrary{decorations.text}
\usetikzlibrary{fadings}
\usetikzlibrary{fit}
\usetikzlibrary{graphs}
\usetikzlibrary{matrix}
\usetikzlibrary{mindmap}
\usetikzlibrary{overlay-beamer-styles}
\usetikzlibrary{patterns}
\usetikzlibrary{positioning}
\usetikzlibrary{scopes}
\usetikzlibrary{shadows}
\usetikzlibrary{shapes.arrows}
\usetikzlibrary{shapes.misc}
\usetikzlibrary{shapes.symbols}
\usetikzlibrary{shapes}
\usetikzlibrary{snakes}
\usetikzlibrary{trees}

\tikzstyle{orfill}=[draw=tblue2!80,top color=tblue2!30,bottom color=tblue2]
\tikzstyle{confl}=[draw=tred2,thick,densely dotted]
\tikzstyle{reqe}=[draw=tblue2!80,thick]
\tikzstyle{pack}=[rectangle,rounded corners=1mm,thick,draw=tblue2!80,%
                  top color=white,bottom color=tblue2!20,%
                  minimum height=.5cm,minimum width=.5cm,inner sep=2pt,draw%
                 ]
\tikzstyle{soln}=[rectangle,rounded corners=1mm,thick,draw=tred2!70,%
                  top color=white,bottom color=tred2!40,%
                  minimum height=.5cm,minimum width=.5cm,inner sep=2pt,draw%
                 ]

\tikzstyle{decision} = [diamond, draw, fill=blue!20,
    text width=4.5em, text badly centered, node distance=3cm, inner sep=0pt]
\tikzstyle{block} = [rectangle, draw, fill=blue!20,
    text width=5em, text centered, rounded corners, minimum height=4em]
\tikzstyle{line} = [draw, -latex']
\tikzstyle{cloud} = [draw, ellipse,fill=red!20, node distance=3cm,
    minimum height=2em]

\tikzset{
    emptyarrowa/.style={
        draw=tred2,
        single arrow,
        minimum height=2.5ex,
        single arrow head extend=1ex
    }
}

\tikzset{
    emptyarrown/.style={
        draw=tgreen3,
        fill=tgreen3,
        single arrow,
        minimum height=2.5ex,
        single arrow head extend=1ex
    }
}

\tikzset{
    invisarrown/.style={
        draw=cover_white,
        fill=cover_white,
        single arrow,
        minimum height=2.5ex,
        single arrow head extend=1ex
    }
}

\definecolor{darkred}{rgb}{0.7,0.1,0.1}
\definecolor{medred}{rgb}{0.5,0.1,0.1}
\definecolor{midred}{rgb}{0.7,0.2,0.2}
\definecolor{vdarkred}{rgb}{0.4,0.1,0.1}
\definecolor{darkslategray}{rgb}{0.18, 0.31, 0.31} 
\definecolor{platinum}{rgb}{0.9, 0.89, 0.89} 
\definecolor{gray}{rgb}{.4,.4,.4}
\definecolor{midgrey}{rgb}{0.5,0.5,0.5}
\definecolor{middarkgrey}{rgb}{0.35,0.35,0.35}
\definecolor{darkgrey}{rgb}{0.3,0.3,0.3}
\definecolor{darkred}{rgb}{0.7,0.1,0.1}
\definecolor{midblue}{rgb}{0.2,0.2,0.7}
\definecolor{darkblue}{rgb}{0.1,0.1,0.5}
\definecolor{darkgreen}{rgb}{0.1,0.5,0.1}
\definecolor{defseagreen}{cmyk}{0.69,0,0.50,0}
\definecolor{purple3}{RGB}{125,38,205}          

\definecolor{tyellow1}{HTML}{FCE94F}
\definecolor{tyellow2}{HTML}{EDD400}
\definecolor{tyellow3}{HTML}{C4A000}
\definecolor{torange1}{HTML}{FCAF3E}
\definecolor{torange2}{HTML}{F57900}
\definecolor{torange3}{HTML}{C35C00}
\definecolor{tbrown1}{HTML}{E9B96E}
\definecolor{tbrown2}{HTML}{C17D11}
\definecolor{tbrown3}{HTML}{8F5902}
\definecolor{tgreen1}{HTML}{8AE234}
\definecolor{tgreen2}{HTML}{73D216}
\definecolor{tgreen3}{HTML}{4E9A06}
\definecolor{tblue1}{HTML}{729FCF}
\definecolor{tblue2}{HTML}{3465A4}
\definecolor{tblue3}{HTML}{204A87}
\definecolor{tpurple1}{HTML}{AD7FA8}
\definecolor{tpurple2}{HTML}{75507B}
\definecolor{tpurple3}{HTML}{5C3566}
\definecolor{tred1}{HTML}{EF2929}
\definecolor{tred2}{HTML}{CC0000}
\definecolor{tred3}{HTML}{A40000}
\definecolor{tlgray1}{HTML}{EEEEEC}
\definecolor{tlgray2}{HTML}{D3D7CF}
\definecolor{tlgray3}{HTML}{BABDB6}
\definecolor{tdgray1}{HTML}{888A85}
\definecolor{tdgray2}{HTML}{555753}
\definecolor{tdgray3}{HTML}{2E3436}

\newcommand{\ncolor}[1]{{\color{tbrown3}#1}}

\newcommand{\fml}[1]{{\mathcal{#1}}}

\newcommand{\tn}[1]{\textnormal{#1}}

\newcommand{\msf}[1]{\ensuremath\mathsf{#1}}
\newcommand{\mbf}[1]{\ensuremath\mathbf{#1}}
\newcommand{\mbb}[1]{\ensuremath\mathbb{#1}}

\newcommand{\waxp}{\ensuremath\mathsf{WAXp}}
\newcommand{\wcxp}{\ensuremath\mathsf{WCXp}}
\newcommand{\axp}{\ensuremath\mathsf{AXp}}
\newcommand{\cxp}{\ensuremath\mathsf{CXp}}
\newcommand{\wcex}{\ensuremath\mathsf{WCEx}}
\newcommand{\cex}{\ensuremath\mathsf{CEx}}

\newcommand{\bcf}{\ensuremath\msf{BCF}}
\newcommand{\bcfn}[1]{\ensuremath\msf{BCF}_{#1}}

\newcommand{\pnode}{\ensuremath\msf{Parent}}
\newcommand{\rnode}{\ensuremath\msf{Root}}
\newcommand{\noterm}{\ensuremath\msf{NonTerminal}}
\newcommand{\isterm}{\ensuremath\msf{Terminal}}
\newcommand{\feat}{\ensuremath\msf{Feature}}
\newcommand{\paths}{\ensuremath\msf{Paths}}
\newcommand{\elit}{\ensuremath\msf{Literal}}
\newcommand{\getlit}{\ensuremath\msf{Literal}}
\newcommand{\allfeat}{\ensuremath\msf{Features}}
\newcommand{\holds}{\ensuremath\msf{Consistent}}
\newcommand{\cls}{\ensuremath\msf{Class}}

\newcommand{\litset}{\ensuremath\msf{Literals}}

\newcommand{\truepath}{\ensuremath\msf{ConsistentPath}}
\newcommand{\iswaxp}{\ensuremath\msf{IsWAXp}}
\newcommand{\issomewaxp}{\ensuremath\msf{IsSomeWAXp}}

\newcommand{\oneaxp}{\ensuremath\msf{FindAXp}}

\newcommand{\pedt}{\ensuremath\msf{PredictivelyEquivalent}}
\newcommand{\predpedt}{\ensuremath\msf{PredEquiv}}

\newcommand{\relopn}[1]{\ensuremath\msf{op}_{#1}}
\newcommand{\rrelopn}[1]{\ensuremath\,\relopn{#1}\,}
\newcommand{\domain}{\ensuremath{\msf{dom}}}
\newcommand{\sympy}{{\sffamily{\smaller[1]{SymPy}}}\xspace} 

\definecolor{gray}{rgb}{.4,.4,.4}
\definecolor{midgrey}{rgb}{0.5,0.5,0.5}
\definecolor{middarkgrey}{rgb}{0.35,0.35,0.35}
\definecolor{darkgrey}{rgb}{0.3,0.3,0.3}
\definecolor{darkred}{rgb}{0.7,0.1,0.1}
\definecolor{midblue}{rgb}{0.2,0.2,0.7}
\definecolor{darkblue}{rgb}{0.1,0.1,0.5}
\definecolor{defseagreen}{cmyk}{0.69,0,0.50,0}
\newcommand{\jnote}[1]{\medskip\noindent$\llbracket$\textcolor{darkred}{joao}: \emph{\textcolor{middarkgrey}{#1}}$\rrbracket$\medskip}

\newcommand{\jnoteF}[1]{}

\newcounter{Comment}[Comment]
\DeclareMathOperator*{\nentails}{\nvDash}
\DeclareMathOperator*{\entails}{\vDash}

\DeclareMathOperator*{\limply}{\rightarrow}

\tikzset{
  0 my edge/.style={densely dashed, my edge},
  my edge/.style={-{Stealth[]}},
}

\hypersetup{
  colorlinks = true,
  linkcolor = MidnightBlue,
  urlcolor  = MidnightBlue,
  citecolor = MidnightBlue,
  anchorcolor = MidnightBlue
}

\setlist{nosep}

\makeatletter
\renewcommand\paragraph{\@startsection{paragraph}{4}{\z@}%
                                      {0.275ex \@plus0.2125ex \@minus.125ex}
                                      {-1em}%
                                      {\normalfont\normalsize\bfseries}}
\makeatother

\begin{document}

\title{Efficient \& Correct Predictive Equivalence for Decision Trees}

\author{%
  \name Joao Marques-Silva \email jpms@icrea.cat \\
  \addr
  ICREA \& University of Lleida\\
  Lleida, Catalunya, Spain
  \AND
  \name Alexey Ignatiev \email alexey.ignatiev@monash.edu \\
  \addr
  Faculty of Information Technology\\
  Monash University\\
  Melbourne, Australia
}

\ShortHeadings{Predictive Equivalence for DTs}{Marques-Silva and Ignatiev}

\editor{TbD}

\maketitle

\begin{abstract}
  The Rashomon set of decision trees (DTs) finds several relevant
  uses.
  Recent work showed that DTs computing the same classification
  function, i.e. predictive equivalent DTs, can represent a
  significant fraction of the Rashomon set. Such redundancy can be
  undesirable. For example, feature importance based on the Rashomon
  set becomes inaccurate due the existence of predictive equivalent
  DTs, i.e.\ DTs that make the same prediction for every possible
  input.
  In recent work, McTavish et al.\ proposed solutions for several
  computational problems related with DTs, including that of deciding
  predictive equivalent DTs.
  %
  The approach of McTavish et al.
  consists of applying the well-known method of Quine-McCluskey (QM)
  for obtaining minimum-size DNF (disjunctive normal form)
  representations of DTs, which are then used for comparing DTs for
  predictive equivalence.
  Furthermore, the minimum-size DNF representation was also applied to
  computing explanations for the predictions made by DTs, and to
  finding predictions in the presence of missing data. 
  However, the problem of 
  formula minimization is hard for the second level of the
  polynomial hierarchy, and there exist different reasons why the QM
  method may exhibit worst-case exponential running time and space.
  This paper first demonstrates that there exist decision trees that
  trigger the worst-case exponential running time and space of the QM
  method.
  Second, the paper shows that the QM method may incorrectly decide
  predictive equivalence, if two key constraints are not respected,
  and one may be difficult to formally guarantee.
  %
  Third, the paper shows that any of the problems to which the
  minimum-size DNF representation has been applied to can in fact be
  solved in polynomial time, in the size of the DT.
  The experiments confirm that, for DTs for which the worst-case
  of the QM method is triggered, the algorithms proposed in
  this paper are orders of magnitude faster than the ones
  proposed by McTavish et al.
  %
  %
  Finally, the paper relates predictive equivalence with logic-based
  explanations and measures of importance.
\end{abstract}


\clearpage \tableofcontents \clearpage

\section{Introduction} \label{sec:intro}

In decision tree learning, the Rashomon
set~\citep{breiman-ss01,rudin-nips22} captures the set of decision
trees (DTs) for a given classification problem which are within some
distance from some accepted definition of optimal
DT~\citep{rudin-nips22}. Several uses of the Rashomon set have been
studied in recent
years~\citep{rudin-jmlr19,rudin-corr19,rudin-ss22,rudin-facct22,rudin-nips22,rudin-nips23,calmon-isit23,biecek-corr23,biecek-ieee-jbhi24,chen-iclr24,gionis-kdd24,barnard-iclr24,barnard-corr24,rudin-cvpr25,hoffman-corr25}.
For the same classification problem, different decision trees capture
different classification functions, and it is often of interest to
study a sample of the decision trees in the Rashomon set.
Recent work~\citep{rudin-icml25,rudin-corr25} studies predictive
equivalent DTs in the Rashomon set, i.e.\ DTs which represent the same
classification function.
This earlier work observes that the predictive equivalent DTs in the
Rashomon set can represent a fairly significant percentage of the DTs,
and so their removal enables keeping in the Rashomon set only DTs that
capture different classification functions, i.e.\ that are not
predictive equivalent. For example, this yields better estimates in
terms of feature importance~\citep{rudin-nips23}.

To identify DTs that are predictive equivalent, McTavish et
al.~\citep{rudin-icml25,rudin-corr25} propose a solution that revisits
work on boolean formula minimization from the 1950s, namely the
well-known Quine-McCluskey (QM) boolean formula minimization
method~\citep{quine-amm52,quine-amm55,mccluskey-bstj56}.%
\footnote{%
There are different algorithms that implement variants of the QM
method~\citep{mccluskey-bstj56,brayton-bk84,coudert-integration94}.
The work of McTavish et al.~\citep{rudin-icml25,rudin-corr25} exploits
the implementation of QM available in \sympy~\citep{meurer-peerj17}.
}
Starting from some DT $\mathcal{T}$,
the QM method is applied to computing a minimum-size boolean formula
$\mathcal{T}_{\mathsf{DNF},c}$ for each class $c$, in disjunctive
normal form (DNF). This formula is then used not only for deciding
predictive equivalence, but also to answer a number of important
queries regarding DT prediction in terms of non-availability of some
features and their explanation~\citep{rudin-icml25,rudin-corr25}.
It should be underlined that the formula $\fml{T}_{\msf{DNF},c}$ is
\emph{not} any DNF representation of a DT for predicting class $c$.
Since the language of DTs is less expressive that that of DNF
formulas~\citep{rivest-ipl76}, then
any DT can be represented by a DNF of size no larger that the DT.
(Thus, this gives an upper bound on the smallest DNF size for
representing a DT.)
For each class $c$, $\fml{T}_{\msf{DNF},c}$ represents a
\emph{minimum-size} DNF formula obtained from the set of prime
implicants for predicting that class. Moreover, $\fml{T}_{\msf{DNF}}$
represents the set of DNFs, one for each class $c$.
%
%
The approach of McTavish et al.\ proposes the use of the QM method,
subject to two constraints, to obtain a minimal representation, which
is then used for answering different queries, including predictive
equivalence.

In practice, the QM method starts by computing all the prime
implicants of the boolean formula $F$, represented as a disjunction of
terms.
The resulting disjunction is often referred to as the Blake Canonical
Form (BCF)~\citep{blake-phd37,brown-bk90} of $F$. For a DT
$\mathcal{T}$ and predicted class $c$, the BCF will be referred to as
$\mathsf{BCF}_{c}(\mathcal{T})$.
%
Afterwards, the QM method enumerates all the minterms of the function
and then it finds a minimum-cost set covering of the sets of terms in
the formula using a (minimum-cost) subset of the prime implicants of
the formula.
The resulting minimum-size DNF formula is denoted as
$\mathcal{T}_{\mathsf{DNF},c}$, which holds true for \emph{any}
partial assignment that is sufficient for predicting class $c$.
%
%
%
Existing practical evidence indicates that the QM method scales
poorly, with a few tens of variables representing what the QM  method
is known to be able to
tackle~\citep{brayton-bk84,coudert-integration94}.
This is explained by the fact that the generation of prime implicants
is worst-case exponential in time and space, and minimum-cost set
covering corresponds to solving an NP-hard optimization problem, on an
intermediate representation that is worst-case exponential on the size
of the input.
The fact that the QM method scales poorly motivated the development of
a number of alternative
methods~\citep{brayton-bk84,rudell-phd89,coudert-dac93,coudert-soda95,ipms-sat15}.
More importantly, the problem of boolean formula minimization is known
to be $\Sigma_2^p$-hard~\citep{umans-focs98,umans-tcad06} and that the
associated decision problem also lies in the second level of the
polynomial hierarchy.
In practical terms, this matches the complexity of deciding quantified
boolean formulas (QBF) with two levels of quantification, which is
known to scale much worse than solving (co)NP-complete decision
problems on formulas~\citep{sat-handbook21}, e.g.\ satisfiability or
tautology.

While it is known that the number of prime implicants can grow
exponentially with the number of variables for arbitrary boolean
formulas~\citep{markowsky-dm78}, this paper shows that this is also the
case with DTs, i.e.\ there exist DTs for which the BCF is worst-case
exponential in the size of the DT.
%
%
The paper then improves on earlier
work~\citep{rudin-icml25,rudin-corr25}, by proving that the computation
of $\mathsf{BCF}_c(\mathcal{T})$ (which is worst-case exponential in
size) and $\mathcal{T}_{\mathsf{DNF},c}$ (which requires solving an
NP-hard optimization problem on a worst-case exponential size
representation) are both \emph{unnecessary}, and can be replaced with
polynomial-time algorithms.
%
%
Concretely, this paper proves that the properties of:
i) \emph{completeness} (Theorem 3.2 in~\citep{rudin-corr25});
ii) \emph{succinctness} (Proposition 3.3 in~\citep{rudin-corr25}); and
iii) the \emph{resolution of predictive equivalence} (Theorem 3.4
in~\citep{rudin-corr25})
can all be solved in polynomial time in the size of the given DT.
Furthermore, our paper underscores the formal correctness of all the
proposed algorithms.

We also note that, in terms of practical implementations, the issues
with McTavish et al.'s approach may extend beyond the inefficiency of
the proposed algorithms~\citep{rudin-icml25,rudin-corr25}.
The use of the QM method for deciding predictive equivalence hinges on
two constraints. First, the set of prime implicants (the Blake Canonical
Form~\cite{blake-phd37,brown-bk90}) must be sorted. Second, the
execution of the set-covering phase of the QM method must be
deterministic.
%
These two conditions ensure that, even though a minimum-size DNF
representation lacks canonicity, predictive equivalence can be
correctly decided, and so are critical for correctness.
Neverthless, even if the underlying algorithms are sequential and do not
exploit non-determinism, it is the case that most programming
languages do not ensure deterministic
execution~\cite{ernst-icse21,wei-icse25}.
%
%
Thus, and to the best of our knowledge, one of the conditions required
my McTavish et al.\ is not formally guaranteed in practice.

We briefly summarize here the main consequences from the potential
unsoundness of using the QM method for predictive equivalence.
For two DTs that are not predictive equivalent, the McTavish et al's
algorithm correctly report that there is no predictive equivalence.
However, there can exist predictive equivalent DTs for
which the an non-deterministic implementation of the QM methods will
produce distinct minimum-size representations. That would cause the
algorithm of McTavish et al.\ to report non-equivalence.

This paper is organized as follows. 
\cref{sec:prelim} introduces the notation and definitions used
throughtout.
\cref{sec:relw} develops an in-depth analysis of the work of McTavish
et al.~\citep{rudin-icml25,rudin-corr25}.
\cref{ssec:mbdsr:perf} analyzes the approach of McTavish et
al.\ approach in terms of running time, proving that the core
algorithm can require exponential time and space, and also
demonstrating that the worst-case running time and space is exercised
in the case of DTs.
\cref{ssec:mbdsr:errors} goes a step further, and shows that the use
of the QM method for deciding predictive equivalence can produce
incorrect results without the constraints imposed by McTavish et al.
Furthermore, it is argued that one of these constraints may be
unrealistic to achive in practice.
\cref{sec:ptime} proposes polynomial-time algorithms for the three
computational problems studied by McTavish et
al.~\citep{rudin-icml25,rudin-corr25}.
\cref{sec:xtra} proves a number of additional results, relating
predictive equivalence with logic-based
explanations~\citep{ms-rw22,msi-aaai22},
but also with measures of importance, including the
Shapley values~\citep{shapley-ctg53,elkind-bk12} and
Banzhaf values~\citep{banzhaf-rlr65,elkind-bk12}. These additional
results also allow developing an algorithm for predictive equivalence
that asymptotically improves on the algorithm proposed
in~\cref{sec:ptime}.
\cref{sec:res} presents simple experiments demonstrating the massive
difference in performance between the approach of McTavish et al.\ and
the algorithms proposed in this paper.
\cref{sec:conc} concludes the paper.

\paragraph{Disclaimer.}
%
Before this manuscript was submitted for publication,
\cite{rudin-icml25,rudin-corr25} had been informed of the efficiency
limitations of their proposed algorithms, and of the possible issues
of soundness resulting from imposing the constraint of deterministic
execution on the set-covering step of the QM method, since this
constraint may be difficult to guarantee in practice~\cite{ernst-icse21}.

\section{Preliminaries} \label{sec:prelim}

\subsection{Machine Learning (ML) Models}

\paragraph{Classification models.}
A classification problem is defined on a set of features
$\fml{F}=\{1,\ldots,m\}$. Each feature $i$ has a domain
$\mbb{D}_i$. Domains can be categorical or ordinal. If ordinal,
domains can be integer or real-valued.
Feature space is defined as the cartesian product of the domains,
$\mbb{F}=\mbb{D}_1\times\cdots\times\mbb{D}_m$.
A classifier computes a classification function $\kappa$, that maps
feature space to a set of classes $\mbb{K}=\{c_1,\ldots,c_K\}$,
i.e.\ $\kappa:\mbb{F}\to\mbb{K}$.%

%
Given a feature $i$, $x_i$ represents a variable that takes values
from $\mbb{D}_i$, whereas $v_i$ represents a constant taken from
$\mbb{D}_i$.
An instance (or sample) is a pair $(\mbf{v},c)$, with
$\mbf{v}\in\mbb{F}$ and $c\in\mbb{K}$, such that $\kappa(\mbf{v})=c$.

A classification model is defined as a tuple
$C=(\fml{F},\mbb{F},\mbb{K},\kappa)$. (The domains of features are
left implicit for simplicity.)

\paragraph{Regression models.}
%
For regression models, a model similar to the classification case is
assumed. Instead of classes we now assume a set of values $\mbb{V}$.
Moreover, the regression function $\rho$ maps feature space to the set
of values, $\rho:\mbb{F}\to\mbb{V}$.
Although the paper focuses solely on classification models, we could
easily adopt a unified ML model with a prediction function $\pi$,
without explicitly distinguishing classification from
regression~\citep{ms-isola24}.

In this general setting, an ML model is defined as a tuple
$M=(\fml{F},\mbb{F},\mbb{T},\pi)$, where $\mbb{T}$ is a set of
targets; these can represent ordinal or categorical values. (As
before, the domains of features are left implicit for simplicity.)

\subsection{Logic Foundations}

\paragraph{Propositional logic.}
Standard notation and definitions are adopted~\citep{sat-handbook21}.
Boolean formulas are defined over propositional atoms (or variables)
using the standard logical operators $\{\land,\lor,\neg\}$. A literal
is an atom ($x$) or its negation ($\neg{x}$). A term $t$ is a
conjunction of literals. A DNF (disjunctive normal form) formula
$\phi$ is a disjunction of terms. Terms and formulas are also
interpreted as functions, mapping feature space to $\{\top,\bot\}$.%
\footnote{%
For simplicity we do not explicitly define the semantics of
propositional logic formulas, since it is quite standard.}

An \emph{implicant} $t$ for a function $\phi$ is a term that entails
$\phi$, i.e.\ $t\entails\phi$, signifying that
$\forall(\mbf{x}\in\mbb{F}).\left(t(\mbf{x})\limply\phi(\mbf{x})\right)$~\citep{brown-bk90,crama-bk11}.%
\footnote{%
In the boolean case, we will treat functions and formulas
indistinctly.}
A \emph{prime implicant} of $\phi$ is an implicant $t$ such that any
term $t'$, obtained by dropping any literal from $t$, is no longer an
implicant of $\phi$. A prime implicant is \emph{essential} if it
covers at least one point of its domain that is not covered by any
other prime implicant. A \emph{minterm} is an implicant consistent
with exactly one point in feature space.
%
%
Prime implicants are often associated with \emph{boolean}
functions~\citep{quine-amm52,quine-amm55,mccluskey-bstj56,brown-bk90,crama-bk11}.
However, prime implicants have been studied in more general
settings~\citep{aiken-cav12}. Similar to more recent
works~\citep{aiken-cav12}, in this paper prime implicants are not
restricted to boolean functions.

\paragraph{Partial assignments \& consistency.}
%
Let $\mbb{OP}$ denote some universe of relational operators, e.g.\ we
could consider $\mbb{OP}=\mbb{OP}_s\cup\mbb{OP}_t$, with
$\mbb{OP}_t=\{\in\}$ and $\mbb{OP}_s=\{=,<,\le,\ge,>\}$.
%
(Logic negation can be used to obtain additional operators.)
Let
$\mbb{L}=\{(x_i,R_i,\relopn{i})\,|\,(i\in\fml{F})\land(R_i\subseteq\mbb{D}_i)\land(\relopn{i}\in\mbb{OP}_t)\}\cup\{(x_i,R_i,\relopn{i})\,|\,(i\in\fml{F})\land(R_i\in\mbb{D}_i)\land(\relopn{i}\in\mbb{OP}_s)\}$
denote all the literals that can be obtained from the variables
associated with the features in $\fml{F}$ and subsets or values from
their domains.
(Throughout the paper, we will simplify the notation. For example,
$(x_i,v_i)$ will be used instead of $(x_i,v_i,=)$, and both represent
the literal $x_i=v_i$. Moreover, $(x_i,v_i,=)$ is equivalent to writing
$(x_i,\{v_i\},\in)$.)
%
%
The set of \emph{partial assignments} is defined by the powerset of 
$\mbb{L}$, i.e. $\mbb{A}=2^{\mbb{L}}$.
%
%
Thus, a partial assignment $\fml{A}\in\mbb{A}$ is a subset of
$\mbb{L}$. 
In the case of complete assignments (or inputs to an ML model), we
require that all features are assigned a single value from their
domain, 
$\mbb{I}=\{\fml{A}\in\mbb{A}\,|\,\forall(i\in\fml{F}).|\{(x_i,R_i,\relopn{i})\in\fml{A}\}|=1\land(x_i,v_i,=)\in\fml{A}\}$.
With each literal $(x_i,R_i,\relopn{i})\in\mbb{L}$ we associate a
domain
$\domain((x_i,R_i,\relopn{i}))=\{\mbf{x}=(x_1,\ldots,x_i,\ldots,x_m)\in\mbb{F}\,|\,x_i\rrelopn{i}{v_i}\}$.
Thus, the domain of a literal is the set of points in feature space
for which the literal holds true.
Similarly, we define the domain of a partial assignment $\fml{A}$,
i.e.\ $\domain(\fml{A})=\cap_{(x_i,v_i,\relopn{i})}\domain(x_i,v_i,\relopn{i})$.
In addition, a partial assignment is said to be consistent, it if
exhibits a non-empty domain:
\[
\holds(\fml{A})~\coloneq [\domain(\fml{A})\not=\emptyset]
\]
By default, we will require partial assignments to be consistent.
Finally, we note that (prime) implicants can also be represented as
partial assignments~\citep{aiken-cav12,amgoud-ijar23}. This provides a
convenient framework for formalizing explanations in DTs.


\subsection{Classification with Decision Trees}

\paragraph{Decision trees.}
A DT is a connected directed acyclic graph $\fml{T}=(V,E)$, with a
single root node $r\in{V}$. $r=\rnode(\fml{T})$ indicates that
$r\in{V}$ is the root of DT $\fml{T}$. The tree nodes $V$ are
partitioned into non-terminal and terminal nodes.
Each non-terminal node $s$ (such that predicate $\noterm(s;\fml{T})$
holds true) is associated with a feature $i$,
i.e. $i=\feat(s;\fml{T})$. For a non-terminal node $s$, associated
with feature $i$, the domain of $i$ is partitioned among the children
of node $s$. Each terminal node $s$ (such that predicate
$\isterm(s;\fml{T})$ holds true) is associated with a class from
$\mbb{K}$, denoted by $\cls(s; \fml{T})$. Whereas non-terminal nodes
have two or more children, terminal nodes have no children. Moreover,
for each node $s$ in the DT, $\pnode(s;\fml{T})$ represents the parent
of $s$ in the tree. For the root node $r=\rnode(\fml{T})$,
$\pnode(r;\fml{T})=\msf{none}$.
As with any classifier, a DT computes a classification function
$\kappa_{\fml{T}}$, mapping feature space to $\mbb{K}$.
For convenience, $\noterm(\fml{T})$ denotes the nodes in $V$ that are
non-terminal. Similarly, $\isterm(\fml{T})$ denotes the nodes in $V$
that are terminal.
%

A path $P$ in a DT is a sequence of nodes
$P=\langle{s_1},\ldots,{s_r}\rangle$, such that $s_1=\rnode(\fml{T})$,
$s_j=\pnode(s_{j+1};\fml{T}),j=1,\ldots,r$, and
$\isterm(s_r,;\fml{T})$ holds true. Moreover, each pair
$(s_j,s_{j+1})$ represents both an edge of $P$ and of $\fml{T}$, 
i.e.\ $(s_j,s_{j+1})\in{E}$.
We will use $\isterm(P;\fml{T})$ to denote the terminal node of
path $P$ in DT $\fml{T}$.
The paths of DT $\fml{T}$ are denoted by $\paths(\fml{T})$.
Moreover, if $(s_p,s_q)$ is an edge of $\fml{T}$, then
$\elit(s_p,s_q;\fml{T})$ represents the literal associated 
with $\feat(s_p;\fml{T})$ and edge $(s_p,s_q)$ on the path of
$\fml{T}$ containing that edge.
For a path $P\in\paths(\fml{T})$,
$\litset(P)=\{\elit(s_p,s_q;\fml{T})\,|\,(s_p,s_q)\in{P}\}$.
We note that $\litset(P)$ represents a generalized partial assignment
and so the above definition of consistency also applies to the
literals in a DT path.

%
%


In this paper, we assume conditions that are often left implicit when
studying DTs, and which can be formalized as follows:
%
%
\begin{align}
  \forall&(P\in\paths(\fml{T})).%
  \left(\litset(P;\fml{T})\nentails\bot\right)\\
  \forall&(\mbf{x}\in\mbb{F}).%
  \exists(P\in\paths(\fml{T})).\left(\mbf{x}\entails\litset(P;\fml{T})\right)
  \\
  \forall&(\mbf{x}\in\mbb{F}).%
  \forall(P\in\paths(\fml{T})).
  \left(\left(\mbf{x}\entails\litset(P;\fml{T})\right)\limply\right.\nonumber\\
  &\left.\left(\forall(Q\in\paths(\fml{T})\setminus\{P\}).\left(\mbf{x}\nentails\litset(Q;\fml{T})\right)\right)\right)
\end{align}
(For simplicity, $\litset$ is interpreted as the conjunction of the
literals.)
Thus, DTs are assumed to respect:
(i) any path in the DT is consistent;
(ii) any point in feature space is consistent with at least one tree
path; and
(iii) any point in feature space that is consistent with some tree
path must not be consistent with any other tree path.

%
%
%
Two DTs $\fml{T}$ and $\fml{T}'$ are \emph{predictive equivalent} if
it holds that:
$\forall(\mbf{x}\in\mbb{F}).\left(\kappa_{\fml{T}}(\mbf{x})=\kappa_{\fml{T}'}(\mbf{x})\right)$.
Thus, predictive equivalence signifies that the two DTs cannot be
distinguished over any point in feature space.
For any pair of DTs $\fml{T}_1,\fml{T}_2$, the predicate
$\predpedt(\fml{T}_1,\fml{T}_2)$ takes value $\top$ (i.e.\ it holds
true) if the pair of DTs is predictive equivalent; otherwise it takes
value $\bot$.
Although the paper focuses mostly on DTs, it is plain that the
definition of predictive equivalence can be used to relate two
arbitrary ML models. For example, we could have a DT that is
predictive equivalent to some random forest (RF)~\citep{breiman-ml01}.
In~\cref{sec:xtra} we revisit predictive equivalence in this more
general setting.

Although it is well-known that learning an optimal decision tree is
NP-hard~\citep{rivest-ipl76}, recent years witnessed a renewed interest
in the learning of optimal decision trees
(e.g.~\citep{nijssen-kdd07,hebrard-cp09,nijssen-dmkd10,bertsimas-ml17,verwer-cpaior17,nipms-ijcai18,verwer-aaai19,rudin-nips19,schaus-cj20,avellaneda-aaai20,schaus-aaai20,rudin-icml20,janota-sat20,hebrard-ijcai20,schaus-ijcai20b,schaus-ijcai20a,demirovic-aaai21,mcilraith-cp21,demirovic-jmlr22,rudin-ss22,demirovic-nips23},
among many others), with the purpose of improving ML model
interpretability.
The rationale is that DTs are \emph{interpretable} and so the model
can be used for obtaining an explanation, e.g.\ the path consistent
with the given instance~\citep{molnar-bk20}.
Moreover, and although it is generally accepted that DTs are
interpretable~\citep{breiman-ss01},%
\footnote{%
The perception of DT interpretability can be traced back more than two 
decades.
A quote from a 2001 paper by L.\ Breiman~\citep{breiman-ss01}
illustrates this perception: \emph{``On interpretability, trees rate
an A+''}.}
there is no rigourous definition of \emph{interpretability} or
\emph{interpretable} ML model and this is unlikely to
change~\citep{lipton-cacm18}.
Furthermore, a number of recent
works~\citep{iims-corr20,iims-jair22,marquis-dke22,msi-frai23,rudin-corr25}  
observe that even optimal decision trees need to be explained,
e.g.\ using a formal definition of
explanation~\citep{ms-rw22,msi-aaai22}.
Concretely, recent work demonstrated that if DTs are explained by
reporting the path consistent with some instance, then the explanation
can be arbitrarily redundant on the number of
features~\citep{iims-corr20,iims-jair22}.
Accordingly, one of the concerns of the approach of McTavish et
al.~\citep{rudin-icml25,rudin-corr25} is to also find rigorous
explanations for DTs.


\begin{figure*}[t]
  \begin{subfigure}[b]{0.325\textwidth}
    \centering
    \scalebox{0.995}{
%
\forestset{
  BDT/.style={
    for tree={
      l=1.5cm,s sep=1.15cm,
      if n children=0{}{circle}, 
      draw=midblue,
      text=midblue,
      edge={
        my edge
      },
      edge=thick,
    }
  },
}
\begin{forest}
  BDT
  [{$x_1$}, label={[yshift=-6.875ex]{{\tiny1}}} 
    [{$x_2$}, label={[yshift=-6.875ex]{{\tiny2}}}, 
      edge label={node[midway,left,xshift=-1.5pt] {{\scriptsize$\in\{0\}$}}}
      [\ncolor{\boldmath{$0$}}, label={[yshift=-5.25ex]{{\tiny4}}},
        edge label={node[midway,left,xshift=-0.5pt]
          {{\scriptsize$\in\{0\}$}}}, rectangle, fill={torange1!10} ]
      [\ncolor{\boldmath{$1$}}, label={[yshift=-5.25ex]{{\tiny5}}},
        edge label={node[midway,right,xshift=-0.575pt] {{\scriptsize$\in\{1\}$}}}, rectangle, fill={torange1!10} ]
    ]
    [\ncolor{\textbf{1}}, label={[yshift=-5.25ex]{{\tiny3}}},
      edge label={node[midway,right,xshift=0.5pt] {{\scriptsize$\in\{1\}$}}},
      rectangle, fill={torange1!10} ]
  ]
\end{forest}
    }
    \caption{Decision tree, $\fml{T}_1$}
  \end{subfigure}
  \begin{subfigure}[b]{0.325\textwidth}
    \centering
    \scalebox{0.995}{
%
\forestset{
  BDT/.style={
    for tree={
      l=1.5cm,s sep=1.15cm,
      if n children=0{}{circle}, 
      draw=midblue,
      text=midblue,
      edge={
        my edge
      },
      edge=thick,
    }
  },
}
\begin{forest}
  BDT
  [{$x_2$}, label={[yshift=-6.875ex]{{\tiny1}}} 
    [{$x_1$}, label={[yshift=-6.875ex]{{\tiny2}}}, 
      edge label={node[midway,left,xshift=-1.5pt] {{\scriptsize$\in\{0\}$}}}
      [\ncolor{\boldmath{$0$}}, label={[yshift=-5.25ex]{{\tiny4}}},
        edge label={node[midway,left,xshift=-0.5pt]
          {{\scriptsize$\in\{0\}$}}}, rectangle, fill={torange1!10} ]
      [\ncolor{\boldmath{$1$}}, label={[yshift=-5.25ex]{{\tiny5}}},
        edge label={node[midway,right,xshift=-0.575pt] {{\scriptsize$\in\{1\}$}}}, rectangle, fill={torange1!10} ]
    ]
    [\ncolor{\textbf{1}}, label={[yshift=-5.25ex]{{\tiny3}}},
      edge label={node[midway,right,xshift=0.5pt] {{\scriptsize$\in\{1\}$}}},
      rectangle, fill={torange1!10} ]
  ]
\end{forest}}
    \caption{DT $\fml{T}_2$, $\predpedt(\fml{T}_1,\fml{T}_2)$}
  \end{subfigure}
  \begin{subfigure}[b]{0.325\textwidth}
    \centering
    \scalebox{0.995}{
%
\forestset{
  BDT/.style={
    for tree={
      l=1.5cm,s sep=1.15cm,
      if n children=0{}{circle}, 
      draw=midblue,
      text=midblue,
      edge={
        my edge
      },
      edge=thick,
    }
  },
}
\begin{forest}
  BDT
  [{$x_2$}, label={[yshift=-6.875ex]{{\tiny1}}} 
    [{$x_1$}, label={[yshift=-6.875ex]{{\tiny2}}}, 
      edge label={node[midway,left,xshift=-1.5pt] {{\scriptsize$\in\{0\}$}}}
      [\ncolor{\boldmath{$1$}}, label={[yshift=-5.25ex]{{\tiny4}}},
        edge label={node[midway,left,xshift=-0.5pt]
          {{\scriptsize$\in\{0\}$}}}, rectangle, fill={torange1!10} ]
      [\ncolor{\boldmath{$0$}}, label={[yshift=-5.25ex]{{\tiny5}}},
        edge label={node[midway,right,xshift=-0.575pt] {{\scriptsize$\in\{1\}$}}}, rectangle, fill={torange1!10} ]
    ]
    [\ncolor{\textbf{1}}, label={[yshift=-5.25ex]{{\tiny3}}},
      edge label={node[midway,right,xshift=0.5pt] {{\scriptsize$\in\{1\}$}}},
      rectangle, fill={torange1!10} ]
  ]
\end{forest}}
    \caption{DT $\fml{T}_3$, $\neg\predpedt(\fml{T}_1,\fml{T}_3)$}
  \end{subfigure}

  \medskip
  
  \begin{subfigure}[b]{\textwidth}
    \centering
    \renewcommand{\arraystretch}{1.075}
    \renewcommand{\tabcolsep}{0.3125em}
    \scalebox{0.995}{
      \begin{tabular}{cc} \toprule
        Domains & $\mbb{D}_1=\mbb{D}_2=\{0,1\}$ \\
        \midrule
        Classes & $\fml{K} = \{0,1\}$ \\
        \midrule
        $\fml{T}$ & $V=\{1,2,3,4,5\}$ \\
        $\rnode(\fml{T})$ & $1$ \\
        $\noterm(\fml{T})$ & $\{1,2\}$ \\
        $\isterm(\fml{T})$ & $\{3,4,5\}$ \\
        %
        $\paths(\fml{T})$ & $\{P_1=\langle1,2,4\rangle,P_2=\langle1,2,5\rangle,P_3=\langle1,3\rangle\}$\\
        \bottomrule
      \end{tabular}
    }
    \medskip
    \caption{Partial information for DTs $\fml{T}_1$, $\fml{T}_2$ and
      $\fml{T}_3$}
  \end{subfigure}
  \caption{Running example DT $\fml{T}_1$, predictive equivalent (PE)
    DT $\fml{T}_2$, and non-PE DT $\fml{T}_3$} \label{fig:runex}
\end{figure*}

\paragraph{Running example.}
The DTs shown in~\cref{fig:runex} will be used as the running example
throughout the paper, with the reference DT being the first one.
%
As an example, for $P_1=\{1,2,4\}$, we have that
$\litset(P_1;\fml{T})=\{(x_1,\{0\},\in),(x_2,\{0\},\in)\}$.

\subsection{Logic-Based Explanations} \label{ssec:lbxp}

Recent years witnessed the emergence of rigorous, logic-based,
explanability~\cite{ms-rw22,msi-aaai22,darwiche-lics23}. This
subsection provides a brief account of the progress in this field of
research.

\paragraph{Abductive explanations (AXps).}
AXps have been proposed in a number of
works~\citep{darwiche-ijcai18,inms-aaai19}, and represent an example
of formally defined explanations. An AXp answers a ``Why (the
prediction)?'' question, and can be interpreted as a logic rule.
In this paper, the notation used builds upon the one
from~\citep{amgoud-ijar23}, since it is preferred for describing the
algorithms proposed later in the paper.
A weak abductive explanation (WAXp) is a partial assignment
$\fml{A}\in\mbb{A}$ that entails the prediction:%
\footnote{%
As noted earlier, the notation used is adapted
from~\citep{amgoud-ijar23}, and it is more complicated than the one
used in other works, e.g.~\citep{ms-rw22}. However, for the purposes
of this paper, the more complicated notation will be helpful.}
\begin{equation} \label{def:waxp}
  \forall(\mbf{x}\in\mbb{F}).\left(\bigwedge\nolimits_{(x_i,R_i,\relopn{i})\in\fml{A}}x_i{\rrelopn{i}}{R_i}\right)\limply\left(\kappa(\mbf{x})=c\right)
\end{equation}
Predicate $\waxp(\fml{A};(\mbf{v},c),\fml{T})$ holds true
when~\eqref{def:waxp} holds.%
\footnote{For predicates, parameterizations are shown after ';', but
will be omitted for simplicity.}
A subset-minimal WAXp is referred to as an abductive explanation (AXp). 
WAXps (resp.~AXps) are also referred to as sufficient (resp.~minimal)
reasons for a prediction~\citep{darwiche-jlli23,darwiche-lics23}.
Moreover, the WAXps for prediction $c\in\mbb{K}$ represent the
implicants of predicate $\pi_c:\mbb{F}\to\{\bot,\top\}$, such that
$\pi_c(\mbf{x})=[\kappa(\mbf{x})=c]$. We will use WAXps for $\kappa$
and class $c$ and the implicants of $\pi_c$ interchangeably.

\begin{example}
  For the DT $\fml{T}_1$ and for class 1, $\{(x_1,0),(x_2,1)\}$ is
  a WAXp. Moreover, $\{(x_2,1)\}$ is an AXp. By inspection, another AXp
  for class 1 is $\{(x_1,1)\}$. Finally, for class 0, the only (W)AXp
  is $\{(x_1,0),(x_2,0\}$.
\end{example}

For a given classification model $M$, we define the set of AXps for an
instance $(\mbf{v},c)$ by 
$\mbb{A}((\mbf{v},c);M)=\{\fml{A}\subseteq\mbb{L}\,|\,\holds(\fml{A})\land\axp(\fml{A};(\mbf{v},c),M)\}$.
Furthermore, we observe that, given the definitions above, the
predicate $\waxp$ does not depend on $\mbf{v}$, only on some partial
assignment (that may be extracted from $\mbf{v}$). As a result, when
convenient, we will use the representation $\waxp(\fml{A};c,M)$ that
holds true for partial assignments such that \eqref{def:waxp} holds,
when $M$ computes classification function $\kappa$ and predicts
$c\in\mbb{K}$. This alternative representation will also be used to
represent AXps. 
Thus, we can now define the set of all partial assignments that
represent (W)AXps for a given prediction $c$:
\begin{align} 
  \mbb{WA}(M,c) &
  =\{\fml{A}\subseteq\mbb{L}\,|\,\holds(\fml{A})\land\waxp(\fml{A};c,M)\}
  \label{eq:allwaxpc}
  \\
  \mbb{A}(M,c) &
  =\{\fml{A}\subseteq\mbb{L}\,|\,\holds(\fml{A})\land\axp(\fml{A};c,M)\}
  \label{eq:allaxpc}
\end{align}
(Since partial assignments are chosen from $\mbb{L}$, we have to
ensure that the partial assignment is consistent.)

Furthermore, each AXp is uniquely defined by a partial assignment
and a prediction. We can then talk about the sets of pairs, of partial
assignments and predictions, such that the partial assignments are
sufficient for some prediction $c$,
%
\begin{align}
  \mbb{WA}(M) & =
  \{(\fml{A},c)\,|\,c\in\mbb{K}\land\fml{A}\subseteq\mbb{L}\land\holds(\fml{A})\land\waxp(\fml{A};c,M)\}
  \nonumber\\
  & = \{(\fml{A},c)\,|\,c\in\mbb{K}\land\fml{A}\in\mbb{WA}(M,c)\}
  \nonumber\\
  \mbb{A}(M) & =
  \{(\fml{A},c)\,|\,c\in\mbb{K}\land\fml{A}\subseteq\mbb{L}\land\holds(\fml{A})\land\axp(\fml{A};c,M)\}
  \nonumber\\
  & = \{(\fml{A},c)\,|\,c\in\mbb{K}\land\fml{A}\in\mbb{A}(M,c)\}
  \nonumber 
\end{align}
Observe that the above definitions allow aggregating the (W)AXps for
all of the classes in $\mbb{K}$.

\paragraph{Contrastive explanations (CXps).}
%
CXps can be defined similarly to AXps~\citep{inams-aiia20,ms-rw22}.
However, given the option to work with partial assignments in this
paper, we would have to adapt the definition of CXps to account for
complete assignments. Since the paper's results do not dependent on
CXps, we opt not to discuss CXps in this paper.

\paragraph{Progress in logic-based XAI.}
Besides the core references above, there has been rapid progress in
logic-based XAI, by a growing number of
researchers~\citep{darwiche-ijcai18,inms-aaai19,inms-nips19,darwiche-ecai20,msgcin-nips20,barcelo-nips20,inams-aiia20,kutyniok-jair21,msgcin-icml21,ims-sat21,ims-ijcai21,kwiatkowska-ijcai21a,hiims-kr21,marquis-kr21,cms-cp21,lorini-clar21,mazure-cikm21,marquis-dke22,iisms-aaai22,hiicams-aaai22,marquis-aaai22,rubin-aaai22,darwiche-aaai22,lorini-wollic22,marquis-ijcai22a,marquis-ijcai22b,amgoud-ijcai22,barcelo-nips22,cms-aij23,ihincms-ijar23,darwiche-jlli23,lorini-jlc23,msi-frai23,yisnms-aaai23,hims-aaai23,marquis-aistats23,hcmpms-tacas23,katz-tacas23,marquis-ijcai23a,hms-ecai23,marquis-ecai23,darwiche-jelia23,ccms-kr23,barrett-nips23,iisms-aaai24,katz-icml24,imms-ecai24,katz-ecai24,ihmpims-kr24,lhms-aaai25,katz-aistats25,bounia-uai25,msllm-ijcai25,iirmss-ijcai25,barcelo-pods25},
among others.
Surveys of some of this past work
include~\citep{msi-aaai22,ms-rw22,darwiche-lics23,ms-iceccs23,ms-isola24}.

\subsection{Additional Concepts}

\paragraph{Complexity classes.}
The paper adopts complexity classes that are standard when studying
the computational complexity of decision problems~\citep{arora-bk09}.
Concretely, the paper considers the complexity classes P, NP, coNP and
$\Sigma_2^p$. The class $\Sigma_2^p$ corresponds to the class
$\tn{NP}^{\tn{NP}}$ i.e.\ the class NP augmented with an oracle for
class NP.
%

\paragraph{Non-determinism.}
Let $\mbb{S}(P,I)$ represent the set of solutions for some
computational problem $P$ on input $I$. For example, in the case of
the QM method, if $P$ represents the QM method, and $I$ denotes some
boolean function, then any minimum-cost DNF would be included in
$\mbb{S}(P)$. 
An algorithm $A$ for problem $P$ on input $I$ implements some form of
non-determinism if multiple executions of $A$ can produce different
elements of $\mbb{S}(P)$. If any execution of algorithm $A$ for
problem $P$ on input $I$ always computes the same solution
$S\in\mbb{S}(P,I)$, then $A$ is deterministic.
Non-determinism may represent one or more of an algorithm's design
decisions, e.g.\ the use of randomization, or it may manifest itself
by the specificities of the programming language used for implementing
the algorithm, e.g.\ the implementation of sorting not being stable,
or the use of hashes returning elements in some arbitrary order.
Most algorithms for solving combinatorial problems implement some sort
of non-determinism~\citep{gomes-aaai98,gomes-jar00,sat-handbook21}.
In addition, the problem of non-determinism in sequential programs is
the subject of continued research~\citep{ernst-icse21,wei-icse25}.

\section{Assessment of McTavish et al.'s Approach} \label{sec:relw}


\subsection{Worst-Case Exponential Time \& Space}
\label{ssec:mbdsr:perf}


This section briefly reviews the recently proposed algorithm for
deciding predictive equivalence~\citep{rudin-icml25,rudin-corr25},
which builds on the well-known method of
Quine-McCluskey~\citep{quine-amm52,quine-amm55,mccluskey-bstj56}.
This algorithm solves the $\Sigma_2^p$-hard problem of boolean formula
minimization~\citep{umans-focs98}, and exhibits well-known worst-case
exponential running time and space.
This section also proves that, even in the case of DTs, there exist
cases of DTs for which the exponential worst-case running time and
space is exercised.

The main goal of~\citep{rudin-icml25,rudin-corr25} is to develop a
representation $\mbb{M}$, which corresponds to $\fml{T}_{\msf{DNF}}$ 
in~\citep{rudin-icml25,rudin-corr25}, and which ensures the properties
of faithfulness, completeness, succinctness and enables deciding
predictive equivalence. Another goal is that ensuring these properties
is computationally efficient in the size of the proposed
representation.
The property of faithfulness signifies that if a prediction is
$c\in\mbb{K}$ for the representation $\mbb{M}$ then $c$ is the correct
prediction for the DT $\fml{T}$. Clearly, the original DT respects
this property.
Completeness means that in the presence of missing data, if all
possible assignments to the features with missing values results in
the same prediction $c\in\mbb{K}$, then the prediction should be $c$.
Succinctness captures the ability to report abductive
explanations~\citep{ms-rw22,msi-aaai22} in the case of
DTs~\citep{iims-corr20,iims-jair22,rudin-icml25,rudin-corr25}.
Polynomial-time algorithms for computing abductive explanations are
described in earlier work~\citep{iims-corr20,iims-jair22}.
Finally, deciding predictive equivalence signifies an algorithm for
declaring that two trees represent the same classification function if
and only if that is indeed the case.

\paragraph{BCF generation step.}
McTavish et al.~\citep{rudin-icml25,rudin-corr25} propose the
enumeration of the DT terms for each class. For the DT
in~\cref{fig:runex}, the terms for prediction 1 are $x_1$ and
$\neg{x_1}\land{x_2}$.
Afterwards, each pair of terms is iteratively selected for a 
consensus operation~\citep{brown-bk90}, i.e.\ one term contains an
atom, the other its negation, and the resulting term omits the
atom. If a non-tautologous new term is obtained, then it is added to
the set of terms.
After all consensus operations are exhausted, i.e.\ no more
non-tautologous consensus operations can be performed, the
operation of \emph{absorption}~\citep{brown-bk90} is applied to remove
terms that contain a superset of the literals of some other term.

\begin{example}
  For the running example, the consensus between the two terms yields
  $x_2$. Afterwards, no more new terms can be obtained by applying the 
  consensus operation. Afterwards, absorption is applied, and so
  $\neg{x_1}\land{x_2}$ is removed because of $x_2$. The resulting set 
  of terms is:  $\bcfn{1}(\fml{T}_1)=\{x_1,x_2\}$.
  Similarly, we obtain
  $\bcfn{0}(\fml{T}_1)=\{\neg{x_1}\land\neg{x_2}\}$.
\end{example}

In the general case, the number of prime implicants of a boolean
formula is exponentially large in the size of the starting
formula~\citep{markowsky-dm78,brown-bk90}.

\paragraph{Set covering step.}
Given the obtained set of terms (i.e.\ the prime implicants for the
predicted class), the QM method then finds a cardinality minimum set
of prime implicants that \emph{covers} (i.e.\ entails) each of the
original terms. For the running example, we keep the set
$\{x_1,x_2\}$.
Set covering is a well-known NP-hard problem~\citep{karp-ccc72}. As a
result, all known exact algorithms are worst-case exponential in the
problem formulation size.
A number of practical optimizations are usually implemented in
practice~\citep{coudert-integration94}, and some are applied to the
covering step of the implementation of the QM method proposed in the
work of McTavish et al.~\citep{rudin-icml25,rudin-corr25}.

For the running example $\fml{T}_1$, the BCFs cannot be further
simplified, and so
$\fml{T}_{1,\msf{DNF},1}=\bcfn{1}(\fml{T}_1)=x_1\lor{x_2}$ and
$\fml{T}_{1,\msf{DNF},0}=\bcfn{0}(\fml{T}_1)=\neg{x_1}\land\neg{x_2}$.

\paragraph{Worst-case running time \& space for DTs.}
Given some DT $\fml{T}$, one key question is the worst-case size of
$\bcfn{c}$ for any $c\in\mbb{K}$. We show that there exist DTs for
which $\bcfn{c}$ is worst-case exponential, confirming worst-case
exponential running time and space in the size of the original DT for
the QM method.

It is known that the number of AXps for a concrete instance
$(\mbf{v},c)$ for a decision tree is exponential in the 
worst-case~\citep[Proposition~1]{marquis-dke22}.
This section proposes a different construction, adapted to the
notation used in this paper.
\jnoteF{Include example DT exercising worst-case BCF.}
%
%
%
%
%
\begin{figure*}[t]
  \scalebox{0.9525}{\tikzset{>=Stealth}
\begin{tikzpicture}[->,%
    EV/.style = {font=\footnotesize},
    node distance={2.0cm}, thin,
    term/.style = {draw, thick, rectangle},
    nterm/.style = {draw, thick, circle,minimum width=0.725cm,minimum height=0.725cm,inner sep=1.0pt},
    nconn/.style = {draw,thick, dashed, circle,minimum width=0.725cm,minimum height=0.725cm,inner sep=1.0pt}
  ]
  \node[nterm] (1)                    {$x_{1}$};
  \node[nterm] (2) [right of=1]       {$x_{2}$};
  \node[nterm] (3) [below of=1]       {$x_{2}$};
  \node[term]  (4) [below of=2]       {1};
  \node[term]  (5) [below left of=3]  {1};
  \node[term]  (6) [below right of=3] {0};
  %
  \node[] (7) at (0.15,-0.5)       {\tiny{1}};
  \node[] (8) at (2.15,-0.5)       {\tiny{2}};
  \node[] (9) at (0.0,-2.5)        {\tiny{3}};
  \node[] (10) at (2.0,-2.4)       {\tiny{4}};
  \node[] (11) at (-1.4,-3.815)    {\tiny{5}};
  \node[] (12) at (1.4,-3.815)     {\tiny{6}};
  \draw[thick] (1) -- node [EV, above]  {\scriptsize{$\in\{1\}$}} (2);
  \draw[thick] (1) -- node [EV, right]  {\scriptsize{$\in\{0\}$}} (3);
  \draw[thick] (3) -- node [EV, left]   {\scriptsize{$\in\{1\}$}} (5);
  \draw[thick] (3) -- node [EV, right]  {\scriptsize{$\in\{0\}$}} (6);
  \draw[thick] (2) -- node [EV, right]  {\scriptsize{$\in\{0\}$}} (4);
  \node[nterm] (13) [right of=2]        {$x_{3}$};
  \node[nterm] (14) [right of=13]       {$x_{4}$};
  \node[nterm] (15) [below of=13]       {$x_{4}$};
  \node[term]  (16) [below of=14]       {1};
  \node[term]  (17) [below left of=15]  {1};
  \node[term]  (18) [below right of=15] {0};
  %
  \node[] (19) at (4.15,-0.5)      {\tiny{7}};
  \node[] (20) at (6.15,-0.5)      {\tiny{8}};
  \node[] (21) at (4.0,-2.5)       {\tiny{9}};
  \node[] (22) at (6.0,-2.4)       {\tiny{10}};
  \node[] (23) at (2.575,-3.815)   {\tiny{11}};
  \node[] (24) at (5.4,-3.815)     {\tiny{12}};
  \draw[thick] (2) -- node [EV, above]   {\scriptsize{$\in\{1\}$}} (13);
  \draw[thick] (13) -- node [EV, above]  {\scriptsize{$\in\{1\}$}} (14);
  \draw[thick] (13) -- node [EV, right]  {\scriptsize{$\in\{0\}$}} (15);
  \draw[thick] (15) -- node [EV, left]   {\scriptsize{$\in\{1\}$}} (17);
  \draw[thick] (15) -- node [EV, right]  {\scriptsize{$\in\{0\}$}} (18);
  \draw[thick] (14) -- node [EV, right]  {\scriptsize{$\in\{0\}$}} (16);
  \node[nterm] (25) [right of=14]       {$x_{5}$};
  \node[nterm] (26) [right of=25]       {$x_{6}$};
  \node[nterm] (27) [below of=25]       {$x_{6}$};
  \node[term]  (28) [below of=26]       {1};
  \node[term]  (29) [below left of=27]  {1};
  \node[term]  (30) [below right of=27] {0};
  %
  \node[] (30a) at (8.175,-0.5)     {\tiny{13}};
  \node[] (30b) at (10.175,-0.5)    {\tiny{14}};
  \node[] (30c) at (8.0,-2.5)       {\tiny{15}};
  \node[] (30d) at (10.0,-2.4)      {\tiny{16}};
  \node[] (30e) at (6.575,-3.815)   {\tiny{17}};
  \node[] (30f) at (9.425,-3.815)   {\tiny{18}};
  \draw[thick] (14) -- node [EV, above]   {\scriptsize{$\in\{1\}$}} (25);
  \draw[thick] (25) -- node [EV, above]   {\scriptsize{$\in\{1\}$}} (26);
  \draw[thick] (25) -- node [EV, right]   {\scriptsize{$\in\{0\}$}} (27);
  \draw[thick] (27) -- node [EV, left]    {\scriptsize{$\in\{1\}$}} (29);
  \draw[thick] (27) -- node [EV, right]   {\scriptsize{$\in\{0\}$}} (30);
  \draw[thick] (26) -- node [EV, right]   {\scriptsize{$\in\{0\}$}} (28);
  \node[nterm] (32) [right of=26]       {$x_{7}$};
  \node[term]  (33) [right of=32]       {1};
  \node[term]  (34) [below of=32]       {0};
  \node[] (34a) at (12.175,-0.5)  {\tiny{19}};
  \node[] (34b) at (14.0,-0.4)    {\tiny{20}};
  \node[] (34c) at (12.0,-2.4)    {\tiny{21}};
  \draw[thick] (26) -- node [EV, above]  {\scriptsize{$\in\{1\}$}} (32);
  \draw[thick] (32) -- node [EV, above]  {\scriptsize{$\in\{1\}$}} (33);
  \draw[thick] (32) -- node [EV, right]  {\scriptsize{$\in\{0\}$}} (34);
\end{tikzpicture} }
  
  \caption{Motivating example. The instance is $((1,1,1,1,1,1,1),1)$.}
  \label{fig:mexmpl}
\end{figure*}
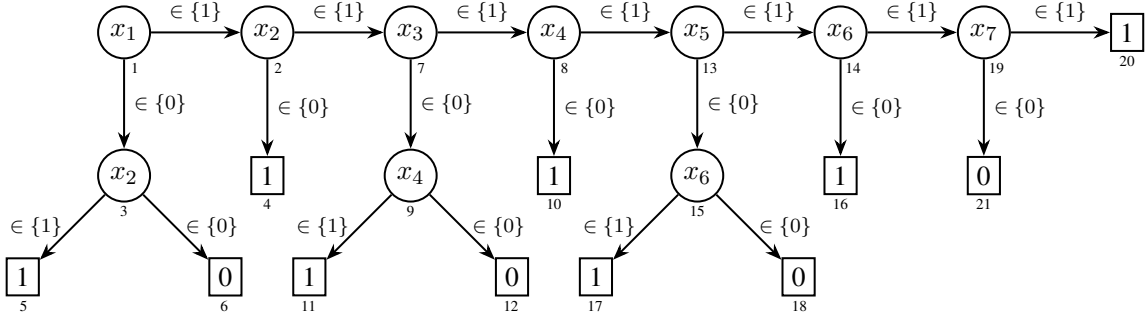
\cref{fig:mexmpl} shows a motivating example for the generic
construction proposed below. All features are binary, and the DT
predicts either class 0 or class 1. The instance is
$((1,1,1,1,1,1,1),1)$.
The prediction can be changed by changing the value of $x_7$, or the
values of both $x_1$ and $x_2$, $x_3$ and $x_4$, or $x_5$ and $x_6$.
Thus, to ensure that the prediction remains unchanged, we must fix
$x_7$, and then fix one of $x_1$ and $x_2$, one of $x_3$ or $x_4$, and
one of $x_5$ or $x_6$.
Thus, the set of AXps for the given instance is 
shown in~\cref{tab:exaxps}.
%
%
\begin{figure}[t]
  \centering
  \begin{tabular}{c} 
    $\begin{aligned}
      \{~~&\{(x_1,1),(x_3,1),(x_5,1),(x_7,1)\},
      ~~\{(x_1,1),(x_3,1),(x_6,1),(x_7,1)\},\nonumber\\
      &\{(x_1,1),(x_4,1),(x_5,1),(x_7,1)\},
      ~~\{(x_1,1),(x_4,1),(x_6,1),(x_7,1)\},\nonumber\\
      &\{(x_2,1),(x_3,1),(x_5,1),(x_7,1)\},
      ~~\{(x_2,1),(x_3,1),(x_6,1),(x_7,1)\},\nonumber\\
      &\{(x_2,1),(x_4,1),(x_5,1),(x_7,1)\},
      ~~\{(x_2,1),(x_4,1),(x_6,1),(x_7,1)\}~~\}\nonumber
    \end{aligned}$
  \end{tabular}
  \smallskip
  \caption{AXps for \cref{fig:mexmpl} and selected instance.}
  \label{tab:exaxps}
\end{figure}
%
There are $8=2^3$ AXps,
resulting from the two options in each of the 
three cases: $x_1$ and $x_2$, $x_3$ and $x_4$, and also $x_5$ and
$x_6$.%
\footnote{%
For simplicity, we opt not to reason in terms of contrastive
explanations~\citep{miller-aij19,inams-aiia20,cms-cp21,cms-aij23},
which would be well-defined in the case of a concrete instance, as is the case in
this example. By using the minimal hitting-set duality between
abductive and contrastive explanations~\citep{inams-aiia20}, it would
also be possible to justify the existence of exponentially many AXps.}
Finally, we note that the example of~\cref{fig:mexmpl} illustrates a
more general construction, which is discussed below.

We now prove the following result.

\begin{theorem} \label{prop:expcase}
  There exist DTs $\fml{T}$ for which $\bcfn{c}(\fml{T})$ is
  exponential in the size of $\fml{T}$ for some class $c$.
  %
\end{theorem}

\begin{figure}[t]
  \begin{subfigure}[b]{0.55\textwidth}
    \centering
    \scalebox{0.995}{\tikzset{>=Stealth}
\begin{tikzpicture}[->,%
    EV/.style = {font=\footnotesize},
    node distance={2.25cm}, thin,
    term/.style = {draw, thick, rectangle},
    nterm/.style = {draw, thick, circle,minimum width=1cm,minimum height=1cm,inner sep=1.0pt},
    nconn/.style = {draw,thick, dashed, circle,minimum width=1cm,minimum height=1cm,inner sep=1.0pt}
  ]
  \node[nterm] (1)                    {$x_{2k-1}$};
  \node[nterm] (2) [right of=1]       {$x_{2k}$};
  \node[nterm] (3) [below of=1]       {$x_{2k}$};
  \node[term]  (4) [below of=2]       {1};
  \node[term]  (5) [below left of=3]  {1};
  \node[term]  (6) [below right of=3] {0};
  \node[nconn] (7) [right of=2]       {?};
  \node[] (8) at (0.35,-0.585) {\tiny{6$k$-5}};
  \node[] (9) at (2.575,-0.585) {\tiny{6$k$-4}};
  \node[] (10) at (0.0,-2.875) {\tiny{6$k$-3}};
  \node[] (11) at (2.265,-2.65) {\tiny{6$k$-2}};
  \node[] (12) at (-1.075,-3.85) {\tiny{6$k$-1}};
  \node[] (13) at (1.175,-3.85) {\tiny{6$k$}};
  %
  \draw[thick] (1) -- node [EV, above]  {\scriptsize{$\in\{1\}$}} (2);
  \draw[thick] (1) -- node [EV, right]   {\scriptsize{$\in\{0\}$}} (3);
  \draw[thick] (3) -- node [EV, left]   {\scriptsize{$\in\{1\}$}} (5);
  \draw[thick] (3) -- node [EV, right]  {\scriptsize{$\in\{0\}$}} (6);
  \draw[thick] (2) -- node [EV, right]   {\scriptsize{$\in\{0\}$}} (4);
  \draw[thick] (2) -- node [EV, above]  {\scriptsize{$\in\{1\}$}} (7);
  %
\end{tikzpicture} }
    \caption{Main gadget}
  \end{subfigure}
  \begin{subfigure}[b]{0.45\textwidth}
    \centering
    \scalebox{0.995}{\tikzset{>=Stealth}
\begin{tikzpicture}[->,%
    EV/.style = {font=\footnotesize},
    node distance={2.25cm}, thin,
    term/.style = {draw, thick, rectangle},
    nterm/.style = {draw, thick, circle,minimum width=1cm,minimum height=1cm,inner sep=1.0pt},
    nconn/.style = {draw,thick, dashed, circle,minimum width=1cm,minimum height=1cm,inner sep=1.0pt}
  ]
  \node[nterm] (1)                    {$x_{2r+1}$};
  \node[term]  (2) [right of=1]       {1};
  \node[term]  (3) [below of=1]       {0};
  \node[] (4) at (0.35,-0.585) {\tiny{6$r$+1}};
  \node[] (5) at (2.25,-0.375) {\tiny{6$r$+2}};
  \node[] (6) at (0.0,-2.65) {\tiny{6$r$+3}};
  \draw[thick] (1) -- node [EV, above]  {\scriptsize{$\in\{1\}$}} (2);
  \draw[thick] (1) -- node [EV, right]   {\scriptsize{$\in\{0\}$}} (3);
\end{tikzpicture} }
    \caption{Final gadget}
  \end{subfigure}
  %
  \caption{Gadgets used in the proof of~\cref{prop:expcase}.
    The main gadget is repeated $r$ times, i.e.\ , $k=1,\ldots,r$.
    The final gadget is connected at the end of the $r$ main gadgets.
  }
  \label{fig:gadgets}
\end{figure}
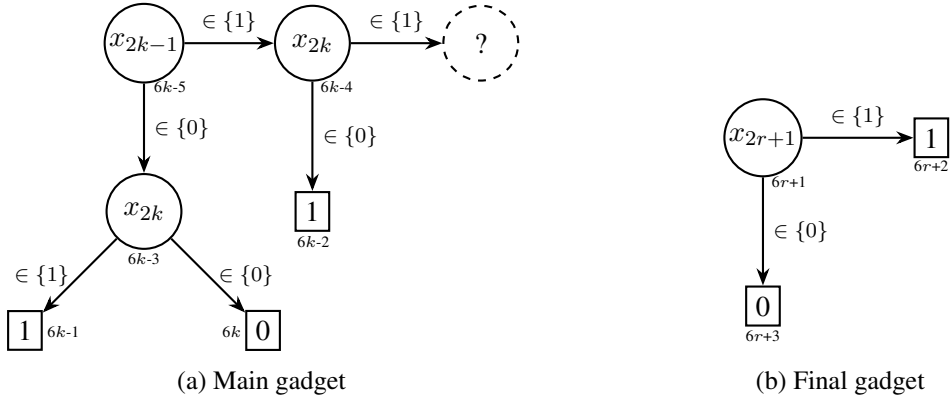

\begin{proof}(Sketch)
  We construct a DT with $2r+1$ features and $6r+3$ nodes, for some
  integer $r$.
  For this DT, a lower bound on the number of AXps is shown to be
  $2^{r}$.
  Concretely, we select an instance, and prove that $2^{r}$
  is a lower bound of the number of AXps, and so on the number of
  prime implicants in the BCF.\\
  The DT is constructed with two gadgets, shown in~\cref{fig:gadgets}.
  The main gadget is replicated $r$ times, as illustrated with the
  example in~\cref{fig:mexmpl}. In addition, the last gadget is linked
  to the last of the $r$ replicas of the main gadget. Clearly, the DT
  size is polynomial on the number of features.
  The chosen instance is $((1,\ldots,1),1)$.
  The rationale for the number of AXps is the same as the one used in
  the analysis of~\cref{fig:mexmpl}.
  There are $r+1$ ways to change the prediction of class 1. One way
  involves changing the value of feature $x_{2r+1}$. Each of the other
  ways involves a pair of variables $x_{2k-1}$ and $x_{2k}$,
  $k=1,\ldots,r$. Thus, in order to ensure that the prediction cannot
  be changed, one must choose from $2^{r}$ possible ways. Each such way
  is subset-minimal, and so it represents an AXp.
  Thus, $\bcfn{1}(\fml{T})$ has worst-case exponential size.
\end{proof}

A more recent report by McTavish et al.~\citep{rudin-corr25v2}
proposes a linear time algorithm for the deciding prediction with
missing data. This algorithm solves a problem already studied in
earlier works, e.g.~\citep[Algorithm~1]{hiims-kr21} but
also~\citep{iims-corr20,iims-jair22}.



\subsection{Potential Unsoundness of QM-based Predictive Equivalence}
\label{ssec:mbdsr:errors}

This section details why the adoption of the QM method for predictive
equivalence can yield incorrect results. \cref{ssec:qminmbdsr} briefly
discusses the constraints imposed in the work of McTavish et al.\ to
circumvent this problem.%
\footnote{%
This discussion is included because the previous work of McTavish et
al.~\citep{rudin-icml25,rudin-corr25} does not discuss the possible
issues resulting from adopting the QM method, neither does it explain
the need for the constraints imposed on the QM method used.}
%

%

In general, if two DTs are not predictive equivalent, then any
implementation of the QM method will correctly report non-equivalence.
However, if two DTs are predictive equivalent, then the QM method is
not in general guaranteed to report equivalence.
%
%
The reason for why the use of the QM method for predictive
equivalence~\citep{quine-amm52,quine-amm55,mccluskey-bstj56} can be
problematic is rather simple.
The Blake canonical form is (as the name indicates)
\emph{canonical}~\citep{blake-phd37,brown-bk90}. Canonicity signifies
that there exists a single unique representation for some boolean
function~\citep{bryant-tcomp86,wegener-bk00,crama-bk11}.
%
However, if a representation is not canonical, then there may exist
\emph{multiple distinct} representations for a given boolean function.
%
A potential problem of adopting the QM methods for predictive
equivalence is that it computes a minimum-cost (e.g.\ number of
literals or terms) representation of some boolean function, and this
is \emph{not} a canonical representation.%
\footnote{Evidently, the concept of canonicity is fundamental for
ensuring uniqueness of
representations~\citep{bryant-tcomp86,wegener-bk00,crama-bk11}.}
%


This section provides additional detail demonstrating the above
arguments, namely the issues with the unrestricted used of the QM
method.
%
%
\cref{ex:error} analyzes an a boolean function with two distinct
minimum cost DNFs. If implementations of QM use any form of
non-determinism, i.e.\ they may not not always report the same
minimum-cost DNF, then the use of QM for predictive equivalence
can declare predictive equivalent DTs not to be equivalent.

\begin{figure}[t]
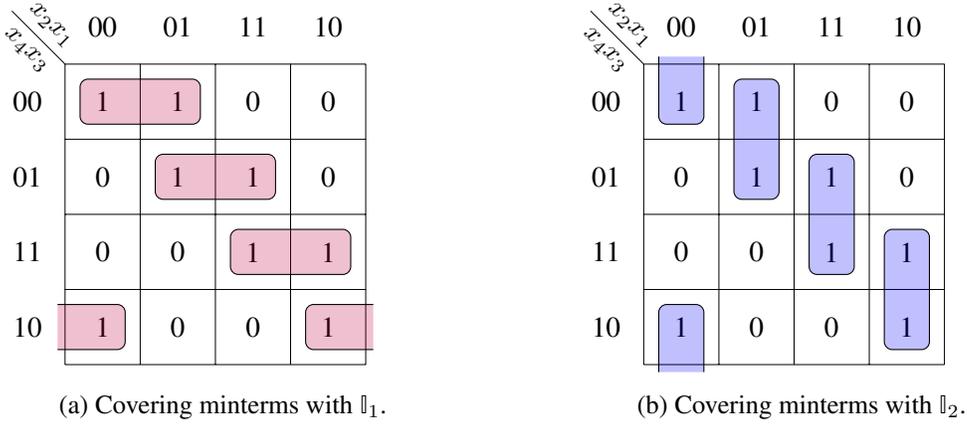

  \begin{subfigure}[b]{0.5\textwidth}
    \centering
    \vspace*{-0.25cm}
    \scalebox{0.9995}{\input{./texfigs/km01a}}
    \vspace*{-0.75cm}
    \caption{Covering minterms with $\mbb{I}_1$.}
  \end{subfigure}
  \begin{subfigure}[b]{0.5\textwidth}
    \centering
    \vspace*{-0.25cm}
    \scalebox{0.9995}{\input{./texfigs/km01b}}
    \vspace*{-0.75cm}
    \caption{Covering minterms with $\mbb{I}_2$.}
  \end{subfigure}
  \caption{Karnaugh maps for boolean function with two distinct
    minimum-cost DNFs.}
  \label{fig:runex02}
\end{figure}

\begin{example} \label{ex:error}
  We consider an example boolean function $f:\{0,1\}^4\to\{0,1\}$,
  with $f(x_4,x_3,x_2,x_1)=\sum_{m_i},i=\{0,2,7,8,10,14,15\}$, where
  each $m_i$ is a minterm.
  \cref{fig:runex02} shows two Karnaugh maps for $f$.%
  \footnote{%
  Karnaugh maps are a well-known method for the simplification of
  boolean functions with a small number of
  variables~\citep{kohavi-bk09}.}
  By inspection of either map, it is plain that the set of prime
  implicants is defined by $\mbb{I}=\mbb{I}_1\cup\mbb{I}_2$, with
  \begin{align}
   \mbb{I}_1 & =
   \{ \bar{x}_4\bar{x}_3\bar{x}_2,
   \bar{x}_4{x}_3{x}_1,
   {x}_4{x}_3{x}_2,
   {x}_4\bar{x}_3\bar{x}_1\} \nonumber \\
   \mbb{I}_2 & =
   \{ \bar{x}_3\bar{x}_2\bar{x}_1,
   \bar{x}_4\bar{x}_2{x}_1,
   {x}_3{x}_2{x}_1,
   {x}_4{x}_2\bar{x}_1
   \} \nonumber
  \end{align}
  Furthermore, again by inspection of the Karnaugh maps, we can
  conclude that both $\mbb{I}_1$ and $\mbb{I}_2$ are minimum-cost DNF
  representations of $f$, as the two maps confirm.
  The set of prime implicants $\mbb{I}_1$ covers all the minterms (in
  red), and the set of prime implicants $\mbb{I}_2$ also covers all
  the minterms (in blue), i.e.\ both are minimum-cost DNFs (for some
  class).
  Either $\mbb{I}_1$ or $\mbb{I}_2$ could be reported by an
  implementation of the QM method.
  This holds true as long as the execution of the QM method is not
  guaranteed to be deterministic.
  %
\end{example}


\subsection{Use of QM in the Approach of McTavish et al.}
\label{ssec:qminmbdsr}


The description of predictive equivalence
in~\cite{rudin-icml25,rudin-corr25} imposes two constraints on the
execution of QM:
(i) the prime implicants must be sorted (and it is implicitly assumed
that the truth table with the minterms is also sorted); and (ii) the
execution of the second phase of the QM method, i.e.\ the set-covering
step, must be deterministic.
\cite{rudin-icml25,rudin-corr25} do not justify the need for these
two constraints.
Nevertheless, with the explicit assumption that these two constraints
hold, then \cite{rudin-icml25,rudin-corr25} prove that predictive
equivalence will be decided correctly.

We underscore that predictive equivalence could be decided correctly
without the second constraint, because BCF is a canonical
representation. Hence, just sorting and comparing the lists of prime 
implicants would suffice to decide predictive equivalence.
The bottom line is that there is not formal need for using the
set-covering step of the QM method and, with that, imposing the need
for deterministic execution. (Evidently, the set-covering step can be
important from the perspective of interpretability.)

Furthermore, ensuring determinism can be a challenge in practice.
This depends on whether or not non-deterministic choices are made by
the set-covering algorithm, but also on whether the programming
language used, and the programming constructs used in the
implementation of the set-covering algorithm, guarantee determinism.
For some programming languages it is known that is \emph{not} the
case~\cite{ernst-icse21,wei-icse25}.
\citep{rudin-icml25,rudin-corr25} do not detail the implementation of
the QM method used, nor discuss whether or not it guarantees
deterministic execution. Analysis of the implementation reveals that
it uses the implementation of the QM method from the \sympy
package~\citep{meurer-peerj17}.
To the best of our knowledge, there exists no formal analysis that the
execution of \sympy is guaranteed to be deterministic.%
\footnote{%
The authors report analyzing the code of \sympy and testing its
results, aiming at ensuring that the execution is
deterministic~\citep{rudin-pc25}.
However, to the best of our knowledge there exist \emph{no} formal
guarantees that the execution of the set-covering step will be
deterministic. As noted earlier, formally ensuring deterministic
execution of sequential programs is a challenge in
itself~\cite{ernst-icse21,wei-icse25}.
}

The previews arguments and the proof
in~\citep[Theorem~3.4]{rudin-icml25,rudin-corr25} justify the following
result. 
\begin{proposition}
  Regarding the use of the QM method for deciding predictive
  equivalence:
  \begin{enumerate}[nosep]
  \item The QM method can yield incorrect results (with respect to
    predictive equivalence) if the input to the set-covering step
    (i.e.\ the sets of prime implicants and minterms) is not sorted
    and the implementation of the QM method is not guaranteed to be
    deterministic.
  \item The QM method will yield correct results (with respect to
    predictive equivalence) if the input to the set-covering step is
    sorted, and the execution of the set-covering step is guaranteed
    to be deterministic. 
  \end{enumerate}
\end{proposition}



Finally, as shown in the next section, the use of the QM method for
deciding predictive equivalence is unnecessary. In fact, this paper
proves that deciding predictive equivalence is polynomial-time
solvable.

\section{Polynomial-Time Algorithms} \label{sec:ptime}

This section shows that the computational overhead of the approach
of~\cite{rudin-icml25,rudin-corr25} can be avoided. Concretely, this
signifies that the computation 
of $\bcf{\fml{T}}$ and $\mathcal{T}_{\mathsf{DNF}}$ (and so the
execution of the QM method) is unnecessary for the computational
problems studied in McTavish et al.'s
work~\citep{rudin-icml25,rudin-corr25}.
To prove this claim, we show that each algorithm proposed
in~\citep{rudin-icml25,rudin-corr25} that uses 
$\mathcal{T}_{\mathsf{DNF}}$ can be replaced by an algorithm with a
running time polynomial in the size of the original DT. 


\paragraph{Basic observations.}
In order to handle missing data, we propose to work with partial
assignments instead of points in feature space. Thus, a feature whose
value is missing is represented by omitting the feature from the
partial assignment.

\jnoteF{\cref{alg:truepath} works for binary features, but it is
  incorrect in general. Why? Because we \textbf{MUST} aggregate all
  the literals on the same feature.}

\begin{algorithm}[t]
  \begin{flushleft}
  \hspace*{\algorithmicindent}
  \textbf{Input}: {$\fml{A}$: partial assignment; $P$: path; $\fml{T}$: DT}\\
  \hspace*{\algorithmicindent}
  \textbf{Output}: {$\top$ if path $P$ consistent with $\fml{A}$;
    else $\bot$}\\
\end{flushleft}

\begin{algorithmic}[1]
  \Function{$\truepath$}{$\fml{A}$, $P$, $\fml{T}$}
  \State{$\forall(i\in\fml{F}).\left(\msf{Lits}[i]\gets\top\right)$}
  \For{$(x_i,R_i,\msf{op}_i)\in\fml{A}$}
  \Comment{Identify features to check for consistency}
  \State{$\msf{Lits}[i]\gets\msf{Lits}[i]\land(x_i\:\msf{op}_i\:R_i)$}
  \EndFor
  \State{$s_b\gets\isterm(P; \fml{T})$}
  \State{$s_a\gets\pnode(s_b; \fml{T})$}
  \While{$s_a\not=\msf{none}$}
  \Comment{Pick literals from tree path, organize by feature}
  \State{$i\gets\feat(s_a)$}
  \State{$\msf{Lits}[i]\gets\msf{Lits}[i]\land\elit(s_a,s_b;\fml{T})$}
  \State{$(s_a,s_b)\gets(\pnode(s_a; \fml{T}), s_a)$}
  \EndWhile
  \For{$i\in\fml{F}$}
  \Comment{Check consistency of the literals for each feature}
  \If{$\neg\holds(\msf{Lits}[i])$}
  \State{\Return $\bot$}
  \Comment{Literals for some feature not consistent}
  \EndIf
  \EndFor
  \State{\Return $\top$}
  \Comment{Literals of every feature in are consistent}
  \EndFunction
\end{algorithmic}

  \caption{Decide consistency of tree path given some partial
    assignment $\fml{A}$}
  \label{alg:truepath}
\end{algorithm}

To decide whether a partial assignment $\fml{A}$ is consistent with a
path $P$, we use~\cref{alg:truepath}.
(It should be noted that~\cref{alg:truepath} serves as a building
block for all the algorithms described in this section.)
For each feature $i\in\fml{F}$, we compute the conjunction of the
literals in $\fml{A}$, if any, with the literals for feature $i$ in
the path $P$, if any. Afterwards, we assess the consistency of the
conjunction of literals of each feature $i$.
The algorithm could be made simpler in the case of binary features,
since there can occur at most one literal on each feature in any
path. However, we opt to describe the general case. Clearly, the
running time of the algorithm is in $\fml{O}(|\fml{F}|+|P|)$, where we
assume that the evaluation of consistency runs in constant time.

\subsection{Completeness \& Succinctness}

\paragraph{WAXp invariance.}
The polynomial-time algorithms described in this section build on the
following result, stating that WAXps are independent of the tree
structure:

\begin{theorem} \label{thm:init}
  Let $\fml{T}$ be some DT.
  For any DT $\fml{T}'$ that is predictive equivalent to $\fml{T}$,
  $\fml{A}\in\mbb{A}$ is a WAXp of $\fml{T}'$ iff $\fml{A}$ is a WAXp
  of $\fml{T}$.
\end{theorem}

The proof idea is that WAXps of some class $c$ are implicants of the
predicate deciding whether the DT predicts $c$. Thus, if the two
DTs compute the same function, then the set of implicants will be the
same. (Observe that the set of implicants is not the BCF of a 
formula, but represents instead the Blake's \emph{complete} canonical 
form~\citep{blake-phd37,brown-bk90}. However, it is also a canonical
representation.)
\smallskip


\begin{proof}(Sketch)
  The WAXps for each class $c$ are implicants for the predicate
  $\pi_c$. Since implicants are a canonical
  representation~\citep{blake-phd37,brown-bk90,darwiche-jair02}, then
  two predictive equivalent DTs will have the same sets of implicants
  for each class. Hence, a partial assignment is an implicant for
  $\pi_c$ if and only if it is an implicant for $\pi'_c$  and
  vice-versa. And also, implicants of $\pi_c$ are WAXps of the
  classifier for class $c$. 
  %
\end{proof}

Given the above, a partial assignment $\fml{A}$ is a WAXp of DT
$\fml{T}$ if and only if it is also a WAXp of any DT $\fml{T}'$ that
is predictive equivalent to $\fml{T}$.
Furthermore, deciding prediction in the presence of missing data is
tightly related with deciding whether a partial assignment represents
a WAXp, i.e.\ it is sufficient for the prediction.


\begin{theorem}
  A partial assignment $\fml{A}$ is a WAXp for class $c$ for DT
  $\fml{T}$ iff for each $\mbf{x}\in\domain(\fml{A})$,
  $\kappa(\mbf{x})=c$. 
  %
\end{theorem}

\begin{proof}
  This result follows from the definition of WAXp.
  %
  %
\end{proof}

\begin{example}
  For the running example DT $\fml{T}_1$, as argued earlier in the
  paper, one of the WAXps is $\{(x_1,0),(x_2,1)\}$. By inspection it
  is clear that the same partial assignment is also a WAXp for
  $\fml{T}_2$ for class 1, but not for $\fml{T}_3$.
\end{example}

Given the above, an algorithm that decides whether a partial
assignment is a WAXp suffices to ensure the property of completeness.


\paragraph{Deciding weak abductive explanations.}
\cref{alg:iswaxp,alg:iswaxpc} summarize the main steps for deciding
whether a partial assignment represents a WAXp for some class
$c\in\mbb{K}$.

\begin{algorithm}[t]
  \begin{flushleft}
  \hspace*{\algorithmicindent}
  \textbf{Input}: {$\fml{A}$: partial assignment; $\fml{T}$: DT}\\
  \hspace*{\algorithmicindent}
  \textbf{Output}: {$\top$ if $\fml{A}$ is WAXp for some class $c$;
    else $\bot$}\\
\end{flushleft}

\begin{algorithmic}[1]
  \Function{$\issomewaxp$}{$\fml{A}$, $\fml{T}$}
  \For{$s\in\isterm(\fml{T})$}
  \Comment{Analyze all terminal nodes}
  \State{$c\gets\cls(s; \fml{T})$}
  \If{$\iswaxp(\fml{A},c,\fml{T})$}
  \Comment{Check if $\fml{A}$ is WAXp for $c$}
  \State{\Return $c$}
  \EndIf
  \EndFor
  \State{\Return $\msf{none}$}
  \EndFunction
\end{algorithmic}

  \caption{Decide whether $\fml{A}$ is a WAXp for some class}
  \label{alg:iswaxp}
\end{algorithm}

\cref{alg:iswaxp} checks whether a partial assignment is a WAXp for 
some class.
Moreover, \cref{alg:iswaxpc} evaluates whether a partial assignment if
a WAXp for a chosen class $c\in\mbb{K}$.
The rationale is that the paths with a prediction other than $c$ are
analyzed for consistency with the partial assignment $\fml{A}$. If
there is consistency, then the partial assignment is not sufficient
for the prediction of $c$; otherwise it is.
Clearly, the algorithms run in polynomial time in the size of a DT.

\begin{algorithm}[t]
  \begin{flushleft}
  \hspace*{\algorithmicindent}
  \textbf{Input}: {$\fml{A}$: partial assignment; $c$: class; $\fml{T}$: DT}\\
  \hspace*{\algorithmicindent}
  \textbf{Output}: {$\top$ if $\fml{A}$ is WAXp for class $c$;
    else $\bot$}\\
\end{flushleft}

\begin{algorithmic}[1]
  \Function{$\iswaxp$}{$\fml{A}$, $c$, $\fml{T}$}
  \For{$P\in\paths(\fml{T})$}
  \Comment{Traverse paths of DT $\fml{T}$}
  \State{$s\gets\isterm(P;\fml{T})$}
  \If{$\cls(s;\fml{T}) \not= c$}
  \Comment{If path prediction differs from goal $c$}
  \If{$\truepath(\fml{A},P,\fml{T})$}
  \State{\Return $\bot$}
  \Comment{$\fml{A}$ consistent with path predicting class other than $c$}
  \EndIf
  \EndIf
  \EndFor
  \State{\Return $\top$}
  \EndFunction
\end{algorithmic}

  \caption{Decide whether $\fml{A}$ is a WAXp for class $c$}
  \label{alg:iswaxpc}
\end{algorithm}

%

\begin{example}
  For the running example DT $\fml{T}_2$, \cref{alg:iswaxpc} can be
  used to decide that $\{(x_1,1)\}$ is a WAXp for class 1.
\end{example}

\paragraph{Computing abductive explanations.}
As noted in recent
work~\citep{iims-jair22,marquis-dke22,rudin-icml25,rudin-corr25},
decision trees can exhibit path literal redundancy, and so the
explanation for a DT prediction can be smaller, possibly much smaller,
than the path size in a  DT.
McTavish et al.'s propose to computing such explanations (and so
deciding predictions given missing data) using $\fml{T}_{\msf{DNF}}$.
However, as proved in~\cref{sec:relw} this is worst-case
exponential in time and space.
In contrast, earlier work proposed a polynomial-time algorithm for
explaining an instance $(\mathbf{v},c)$
(e.g.~\citep{iims-corr20,iims-jair22}).
Here, we opt instead for explaining a partial assignment that is known
to be a WAXp for some class $c$.
\cref{alg:oneaxp} summarizes the computation of one AXp, which
requires that the starting partial assignment is an WAXp for class
$c$.

\begin{algorithm}[t]
  \begin{flushleft}
  \hspace*{\algorithmicindent}
  \textbf{Input}: {$\fml{A}$: partial assignment; $c$: class; $\fml{T}$: DT}\\
  \hspace*{\algorithmicindent}
  \textbf{Output}: {AXp $\fml{X}$ for class $c$ of DT $\fml{T}$}\\
\end{flushleft}

\begin{algorithmic}[1]
  \Function{$\oneaxp$}{$\fml{A}$, $c$, $\fml{T}$}
  \Comment{Precondition: $\waxp(\fml{A},c;\fml{T})$}
  \State{$\fml{X}\gets\fml{A}$}
  \State{$\fml{W}\gets\allfeat(\fml{A})$}
  \For{$i\in\fml{W}$}
  \Comment{Invariant: $\waxp(\fml{X},c;\fml{T})$}
  \State{$(x_i,R_i,\msf{op}_i)\gets\getlit(i,\fml{X})$}
  \State{$\fml{X}\gets\fml{X}\setminus\{(x_i,R_i,\msf{op}_i)\}$}
  \If{$\neg\iswaxp(\fml{X},c,\fml{T})$}
  \State{$\fml{X}\gets\fml{X}\cup\{(x_i,R_i,\msf{op}_i)\}$}
  \EndIf
  \EndFor
  \State{\Return $\fml{X}$}
  \Comment{Final result: $\axp(\fml{X},c;\fml{T})$}
  \EndFunction
\end{algorithmic}

  \caption{Find one AXp}
  \label{alg:oneaxp}
\end{algorithm}

The algorithm's invariant is that the working partial assignment
$\fml{X}$ is a WAXp, from which features are removed while the
invariant is preserved. Due to monotonicity of the definition of WAXp,
the partial assignment that is returned represents an AXp.
Clearly, the algorithm ensures the property of succinctness of
explanations, and runs in polynomial time in the size of the DT.
\cref{alg:oneaxp} mimics the algorithm proposed in earlier
work~\citep{iims-jair22}, but adapted to working with partial
assignments instead of a concrete instance.


\begin{example}
  \cref{alg:oneaxp} can be used to prove that $\{(x_1,1)\}$ is an AXp
  for either $\fml{T}_1$ or $\fml{T}_2$. This is not surprising,
  because the two DTs are predictive equivalent.
  Similarly, starting from $\{(x_1,0),(x_2,1)\}$, \cref{alg:oneaxp}
  will yield $\{(x_2,1)\}$ as the AXp, independently of which DT
  $\fml{T}_1$ or $\fml{T}_2$ is used.
\end{example}

Although not studied in McTavish et al.'s work, one could also compute
contrastive explanations for DTs in polynomial
time~\citep{hiims-kr21,iims-jair22}. For example, the algorithms
detailed in this section could be used with minor changes.

Finally, this section shows that the properties of completeness and
succinctness are tightly related, and in practice are solved with very
similar algorithms, all of which are based on finding/deciding
(W)AXps.

\subsection{Predictive Equivalence} \label{ssec:ptime:pedt}

\paragraph{Deciding predictive equivalence.}
The main goal of the work of McTavish et
al.~\citep{rudin-icml25,rudin-corr25} is to be able to deciding
predictive equivalence efficiently. 
Accordingly, the algorithm proposed in earlier
work~\citep{rudin-icml25,rudin-corr25} runs in polynomial-time in the
size of the $\fml{T}_{\msf{DNF}}$ representation of the DT.
%
%

\begin{theorem} \label{prop:onepoint}
  Two DTs, $\fml{T}_1$ and $\fml{T}_2$,  are not predictively
  equivalent if there exists a point in feature space for which
  $\fml{T}_1$ and $\fml{T}_2$ predict different classes.
\end{theorem}

\begin{proof}
  This result is a consequence of
  the definition of
  predictive equivalence.
\end{proof}

\begin{algorithm}[t]
  \begin{flushleft}
  \hspace*{\algorithmicindent}
  \textbf{Input}: {$\fml{T}_1$: DT 1; $\fml{T}_2$: DT 2}\\
  \hspace*{\algorithmicindent}
  \textbf{Output}: {$\top$ if DTs are predictive equivalent;
    else $\bot$}\\
\end{flushleft}

\begin{algorithmic}[1]
  \Function{$\pedt$}{$\fml{T}_1$, $\fml{T}_2$}
  \For{$P_1\in\paths(\fml{T}_1)$}
  \Comment{Analyze the paths of $\fml{T}_1$}
  \State{$s_1\gets\isterm(P_1;\fml{T}_1)$}
  \State{$c_1\gets\cls(s_1;\fml{T}_1)$}
  \State{$\fml{A}_1\gets\litset(P_1; \fml{T}_1)$}
  \Comment{Create partial assignment}
  \For{$P_2\in\paths(\fml{T}_2)$}
  \Comment{Analyzee the paths of $\fml{T}_2$}
  \State{$s_2\gets\isterm(P_2;\fml{T}_2)$}
  \State{$c_2\gets\cls(s_2;\fml{T}_2)$}
  \If{$c_1\not=c_2$}
  \Comment{Compare the paths if predictions differ}
  \If{$\truepath(\fml{A}_1,P_2,\fml{T})$}
  \State{\Return $\bot$}
  \Comment{Not PE if partial assignment (of $P_1$) consistent with
    path $P_2$}
  \EndIf
  \EndIf
  \EndFor
  \EndFor
  \State{\Return $\top$}
  \Comment{Unable to prove not PE; thus it is PE}
  \EndFunction
\end{algorithmic}

  \caption{Decide predictive equivalence}
  \label{alg:pedt}
\end{algorithm}

From~\cref{prop:onepoint}, it is immediate to conclude that deciding
predictive equivalence is in coNP, and so can be decided with an NP
oracle. (This also means that using the QM method for deciding
predictive equivalence is unnecessary.)
However, we show that deciding predictive equivalence is actually in
P.
%
The rationale of the algorithm is as follows. The algorithm visits the
paths of one of the DTs; we will pick $\fml{T}_1$. For each path $P_1$
with prediction $c$ in $\fml{T}_1$, we create a partial assignment
$\fml{A}_1$ using the literals from the path $P_1$. Afterwards, we
analyze the paths of $\fml{T}_2$ with a prediction \emph{other} than
$c$. If $\fml{A}_1$ is consistent with any such path,
then~\cref{prop:onepoint} applies, and so the two DTs are not
predictively equivalent. 
\cref{alg:pedt} summarizes the main steps of the polynomial-time
procedure to decide predictive equivalence.%
\footnote{%
A different polynomial-time algorithm, restricted to two classes and
binary branching, was proposed in an unpublished 1998 technical
report~\citep{zantema-tr98}.
}
It is plain that the running time is polynomial on the product of the
two DTs. By inspection, a simple upper bound on the running time is  
$\fml{O}(|\fml{T}_1|\times|\fml{T}_2|\times\min(|\fml{T}_1|,|\fml{T}_2|))$.%
\footnote{%
To attain this run-time complexity, the execution of~\cref{alg:pedt}
should be changed to start by comparing the sizes of the two DTs, with
the external loop being run on the largest, and the internal loop
being run on the smallest.}

\begin{figure}
  \begin{subfigure}[b]{0.5\textwidth}
    \centering
    \renewcommand{\tabcolsep}{0.575em}
    \renewcommand{\arraystretch}{1.5}
    \begin{tabular}{ccccc}
      \toprule
      & \multicolumn{1}{c|}{~} & \multicolumn{3}{c}{$\fml{T}_2$} \\
      \cline{3-5}
      & \multicolumn{1}{c|}{~} & $P_1$ & $P_2$ & $P_3$ \\
      \cline{1-5}
      \multirow{3}{*}{$\fml{T}_1$}
      & \multicolumn{1}{|c|}{$P_1$} & --- & $\bot$ & $\bot$ \\
      & \multicolumn{1}{|c|}{$P_2$} & $\bot$ & --- & ---   \\
      & \multicolumn{1}{|c|}{$P_3$} & $\bot$ & --- & ---   \\
      \bottomrule
    \end{tabular}
    \caption{$\domain(\litset(P_i))\cap\domain(\litset(P_j))\not=\emptyset$?}
  \end{subfigure}
  \begin{subfigure}[b]{0.5\textwidth}
    \centering
    \renewcommand{\tabcolsep}{0.575em}
    \renewcommand{\arraystretch}{1.5}
    \begin{tabular}{ccccc}
      \toprule
      & \multicolumn{1}{c|}{~} & \multicolumn{3}{c}{$\fml{T}_3$} \\
      \cline{3-5}
      & \multicolumn{1}{c|}{~} & $P_1$ & $P_2$ & $P_3$ \\
      \cline{1-5}
      \multirow{3}{*}{$\fml{T}_1$}
      & \multicolumn{1}{|c|}{$P_1$} & \color{red}{$\top$} & --- & $\bot$ \\
      & \multicolumn{1}{|c|}{$P_2$} & --- & \color{red}{$\top$} & ---   \\
      & \multicolumn{1}{|c|}{$P_3$} & --- & $\bot$ & ---   \\
      \bottomrule
    \end{tabular}
    \caption{$\domain(\litset(P_i))\cap\domain(\litset(P_j))\not=\emptyset$?}
    \label{fig:pedt:tab2}
  \end{subfigure}
  \caption{Execution of \cref{alg:pedt} on running
    example. $\fml{T}_1$ and $\fml{T}_2$ are predictive equivalent,
    whereas $\fml{T}_1$ and $\fml{T}_3$ are not.}
  \label{fig:pedt}
\end{figure}

\begin{example}
  \cref{fig:pedt} summarizes the execution of~\cref{alg:pedt} on the
  running example DTs. As can be concluded, the algorithm reports that
  $\fml{T}_1$ and $\fml{T}_2$ are predictive equivalent, since no pair
  of paths (one from $\fml{T}_1$ and one from $\fml{T}_2$) with
  different predictions is consistent. In contrast, $\fml{T}_1$ and
  $\fml{T}_3$ are not predictive equivalent, since there exists at
  least one pair of paths (one from $\fml{T}_1$ and one from
  $\fml{T}_2$) with different predictions that is consistent.
  \cref{fig:pedt:tab2} shows the two examples $(P_1,P_1)$ and
  $(P_2,P_2)$, with the first path taken from one of the trees
  (e.g.\ $\fml{T}_1$), and the second path taken from the other tree
  (e.g.\ resp.~$\fml{T}_3$).
\end{example}

Finally, it is important to observe that~\cref{alg:pedt} can be
parallelized.
Given enough processors, both loops can be parallelized, and so the
total run time, if enough processors are available, can be reduced to
$\fml{O}(\max(|\fml{T}_1|,|\fml{T}_2|))$.
%
This complexity assumes that parallelization can be achieved in two
phases, one for the first loop, and then another phase for the
$\fml{O}(|\fml{T}_1|)$ executions of the second loop. For the second
phase, parallelization of the second loop ensures that checking
consistency between each pair of paths is done in parallel requiring at
most $\fml{O}(\max(|\fml{T}_1|,|\fml{T}_2|))$ time.

\paragraph{Discussion.}
One might argue that the algorithms proposed in this section
have complexity higher than those proposed in earlier
work~\citep{rudin-icml25,rudin-corr25}.
However, the algorithms described in this section run in
polynomial-time in the size of the DT, whereas the algorithms proposed 
in~\citep{rudin-icml25,rudin-corr25} require the prior computation of
both $\bcf(\fml{T})$ and $\fml{T}_{\msf{DNF}}$, and $\bcf(\fml{T})$
was shown to be worst-case exponential in the size of the original DT.
%
Besides larger running times, the algorithms proposed
elsewhere~\citep{rudin-icml25,rudin-corr25} face a hard practical
obstacle when $\bcf(\fml{T})$ is exponentially large, since the
representation size can exceed the available computing resources.

\section{Experimental Evidence}
\label{sec:res}

The work of McTavish et al.~\citep{rudin-icml25,rudin-corr25} offers
comprehensive evidence regarding the importance of eliminating
predictive equivalent DTs from the Rashomon set.
Also, the algorithms proposed in the previous section exhibit
negligible running times, even for DTs of fairly large size. (Our
results concur with similar results were obtained in earlier work when
computing instance-specific explanations~\citep{iims-corr20,iims-jair22}.)
As a result, the focus of our experiments is to demonstrate the
practical effect of exercising the worst-case running time of the QM
method.
%
Since the proof~\cref{prop:expcase} only considers a single instance,
the experiments target the complete compilation of $\bcf$
for different values of $r$.

\paragraph{Experimental setup.}
The algorithm described in~\citep{rudin-corr25} uses
\sympy's implementation of QM~\citep{meurer-peerj17}.%
\footnote{\url{https://www.sympy.org/}.}
We also implemented a prototype in Python, that mimics a standard
implementation of the prime implicant generation step of the QM
method, i.e.\ BCF generation.
The experiments were run on a Macbook with an M3 Pro processor. For
the DT used in the proof of~\cref{prop:expcase}, different values of
$r$ were considered. The parallelization of~\cref{alg:pedt},
summarized in~\cref{ssec:ptime:pedt}, was not considered in a
large-scale cluster; this would be simple to do, for example with
Apache Spark.%
\footnote{\url{https://spark.apache.org/}.}

\begin{table}[t]
  \centering
  \renewcommand{\tabcolsep}{0.5em}
  \begin{tabular}{crrrr} \toprule
    $r$ &
    \multicolumn{1}{c}{\sympy} &
    \multicolumn{1}{c}{$|\bcfn{0}(\fml{T})|$} &
    \multicolumn{1}{c}{$|\bcfn{1}(\fml{T})|$} &
    \multicolumn{1}{c}{BCF time} \\
    \toprule
    3 & 0.13    & 4 & 22 & 0.01 \\
    4 & 0.57    & 5 & 46 & 0.07 \\
    5 & 39.60   & 6 & 94 & 0.84 \\
    6 & 2789.45 & 7 & 190 & 11.28 \\
    7 & ${>}$150000.00 & 8 & 382 & 161.25 \\
    8 & -- --    & 9 & 766 & 2264.62 \\
    9 & -- --    & 10 & 1534 & 64458.55 \\
    \bottomrule
  \end{tabular}
  \caption{Results for different values of $r$,
    $r\le9$. \sympy's QM stopped for $r=7$ after
    150000s. Running times in seconds.
    The number of DT nodes is: $6\times{r}+3$ and the number of
    features is $2\times{r}+1$. } \label{tab:res1}
\end{table}

\begin{table}[t]
  \centering
  \renewcommand{\tabcolsep}{0.75em}
  \begin{tabular}{rrrrrrr} \toprule
    \multicolumn{1}{c}{$r$} &
    \multicolumn{1}{c}{\# DT nodes} &
    \multicolumn{1}{c}{\# features} &
    \multicolumn{1}{c}{$|\bcfn{1}(\fml{T})|$} &
    \multicolumn{1}{c}{One AXp} &
    \multicolumn{1}{c}{isWAXp?} &
    \multicolumn{1}{c}{PE?} \\
    \toprule
    200 & 1203 & 401 & $2^{200}$ & 1.71s & 0.005s & 3.7s \\
    500 & 3003 & 1001 & $2^{500}$ & 26.98s & 0.032s & 57.1s \\
    1000 & 6003 & 2001 & $2^{1000}$ & 224.62s & 0.126s & 469.0s \\
    \bottomrule
  \end{tabular}
  \caption{Running times for DTs much larger than those
    in~\cref{tab:res1}, with $r\in\{200,500,1000\}$.}
  \label{tab:res2} 
\end{table}

\paragraph{Results.}
\cref{tab:res1} summarizes the results for values of
$r$ ranging from 3 to 9, i.e.\ DTs with up to $6\times{r}+3=57$ nodes
(for $r=9$).
%
As can be observed, for class 1 (resp.~class 0), the size of
$|\bcfn{1}(\fml{T})|$ (resp.~$|\bcfn{0}(\fml{T})|$) grows 
exponentially (resp.~linearly) with the value of $r$.
\sympy's QM algorithm shows large running times, and
it was stopped for $r=7$, i.e.~DT with 45 nodes. For the independent
generation of BCF, the running times also grow exponentially (as
expected). For $r=9$, the running times for computing the BCF's are
already massive.
The inefficiency observed is attributed to the fact that Python was
used and, more importantly and as noted earlier in the paper, the
computational overhead of generating the BCF. This overhead is
significant, and worsens for trees with more AXps.
For the algorithms proposed in this paper, we consider the cases
$r=200,500,1000$, as shown in~\ref{tab:res2}. For the case $r=200$,
this represents a DT with 
$6\times{r}+3=1203$ nodes, as shown, and so a DT that is almost 20
times larger than the ones considered for computing the BCF and
running QM). In this case, we have $2\times{r}+1=401$ features
, which makes the use of the QM method impractical.
For this DT and prediction 1, and from the
proof of~\cref{prop:expcase}, we have a lower bound of $2^{200}$ prime
implicants for $\pi_1$.
The computation of one AXp for a partial assignment consistent with
the longest tree path takes 1.71s. Deciding if the same partial
assignment is a WAXp takes 5ms. Moreover, deciding predictive
equivalence of a tree with the same size, but with one terminal node
with a changed prediction, takes 3.7s. These running times are 
negligible when compared with the first (or both) phase(s) of the QM
method on a DT with 57 nodes (for $r=9$), i.e.\ a DT that is less than
a tenth of the case $r=200$.
%
%
The cases $r=500$ and $r=1000$ represent much more complex DTs, with
far more nodes and features. Concretely, for $r=500$ (resp.~$r=1000$),
a lower bound on the number of prime implicants is $2^{500}$
(resp.~$2^{1000}$). The running times for computing one AXp
and for deciding predictive equivalence increase noticeably. However,
as noted in~\cref{sec:ptime},~\cref{alg:pedt} can be parallelized, and
so, on a cluster, the observed run times can be significantly reduced.
Furthermore, the use of a programming language other than Python,
e.g.\ C/C++, would enable significant reductions in the running times.

Finally, we underscore that the experimental results consider a
special DT construction aiming at highlighting the performance
limitations of using an implementation of the QM method. However, DTs
of large size can occur in some high-risk~\citep{rudin-naturemi19}
application domains, where high accuracy is
paramount. For example, \citep{ghiasi-cmpb20} reports a DT with 161
nodes and 20 real-valued features.
In those cases, the use of the QM method will also not scale, even if
high-performance solutions are exploited~\citep{coudert-integration94}
(and these might not ensure deterministic execution).
(We note that \sympy is known to be inefficient for
formulas with more than 8 boolean variables.%
\footnote{%
This is discussed for example
in~\url{https://docs.sympy.org/latest/modules/logic.html}.})

\section{Additional Results} \label{sec:xtra}

This section relates predictive equivalence with several concepts
studied in logic-based XAI.

\subsection{Predictive Equivalence and AXps} 

\begin{theorem} \label{prop:pe-xp}
  Let $C_1=(\fml{F},\mbb{F},\mbb{K},\kappa_1)$ and
  $C_2=(\fml{F},\mbb{F},\mbb{K},\kappa_2)$ be two classifiers.
  Then, $C_1$ is predictive equivalent to $C_2$ if and only if
  for every instance $(\mbf{v},c)$, with $\kappa(\mbf{v})=c$,
  $\mbb{A}(C_1,c))=\mbb{A}(C_2,c))$.
\end{theorem}

\begin{proof}
  From~\cref{ssec:lbxp}, it is the case that $\mbb{A}(M,c)$ represents
  all the prime implicants of $\pi_c$. Thus, $\mbb{A}(M,c)$
  represents the BCF for $\pi_c$. The BCF is a canonical (i.e.\ unique)
  representation for a boolean
  function~\citep{blake-phd37,quine-amm52,brown-bk90}. Hence, for
  predictive equivalent classifiers, and for each $c\in\mbb{K}$,
  only one BCF exists.
  The same applies to every $c\in\mbb{K}$.
  Thus, the result follows.
  %
  %
\end{proof}

\begin{remark}
  \cref{prop:pe-xp} also holds if AXps are replaced by weak AXps.
\end{remark}

\begin{remark}
  \cref{prop:pe-xp} can be generalized to arbitrary ML models,
  e.g.\ regression models, that compute some prediction function.
\end{remark}

Although revealing interesting connections between predictive
equivalence and the sets of AXps/CXps, it may also seem unclear how
\cref{prop:pe-xp} could be useful in practice. 

We illustrate one example use of the connection between predictive
equivalence and (W)AXps.

\begin{theorem}
  Suppose we have two ML models $M_1$ and $M_2$, and two partial 
  assignments $\fml{A}_1$ and $\fml{A}_2$, representing (W)AXps for
  classes $c_1\not=c_2$, respectively for $M_1$ and $M_2$.
  It is the case that,
  if $\holds(\fml{A}_1,\fml{A}_2)$,
  then $M_1$ and $M_2$ are not predictive equivalent.
\end{theorem}

\subsection{Predictive Equivalence and Corrected Measures of Importance}

It is well-known that SHAP scores approximate a theoretical
definition of Shapley values for
XAI~\citep{kononenko-jmlr10,kononenko-kis14}.
In addition, the quality of the approximation provided by SHAP scores
depends on a number of factors.
However, recent work proved that the proposed definition of Shapley
values for
XAI~\citep{kononenko-jmlr10,kononenko-kis14,lundberg-nips17} can
mislead human decision makers~\citep{msh-cacm24,hms-ijar24}, by
assigning no importance to features that are critical for a given
prediction, and by assigning some importance to features that bear no
influence in that prediction.
Since it is unclear when Shapley values for XAI (and so SHAP scores)
can mislead, the use of the existing definition of Shapley values for
XAI should be discouraged.

Recent works proposed different solutions to this
flaw~\citep{izza-aaai24,ignatiev-sat24,lhms-aaai25}.
Of these, we study the one proposed by L\'{e}toff\'{e} et
al.~\citep{lhms-aaai25}, since it can also be based on Shapley values.

The key issue with past definitions of Shapley values for XAI is the
characteristic function used, which is based on the expected value of
the classifier.
The proposed solution for this flaw is to adopt a different
characteristic function that is defined as follows, where
$\fml{S}\subseteq\fml{F}$:
\begin{equation} \label{eq:ccf}
  \upsilon(\fml{S};(\mbf{v},c),M)=
  \left\{
  \begin{array}{lcl}
    1 & \quad & \tn{if $\waxp(\fml{S};(\mbf{v},c),M)$}\\[3pt]
    0 & \quad & \tn{otherwise}\\
  \end{array}
  \right.
\end{equation}
(As shown, the definition of $\upsilon$ is parameterized
on the given instance $(\mbf{v},c)$ and the ML model $M$.)
Given this modified characteristic function, the issues reported with
Shapley values for XAI are eliminated~\citep{lhms-aaai25}.
We refer to the definition of Shapley values using the characteristic
function above as \emph{corrected SHAP scores}.

We can now relate predictive equivalence with corrected SHAP scores.

\begin{theorem} \label{prop:pe-sv}
  Let $C_1,C_2$ be two classifiers defined on the same features, with
  the same feature space, and mapping to the same set of classes.
  Then, if $C_1$ and $C_2$ are predictive equivalent then, for
  any instance, the corrected SHAP scores for $C_1$ equal the
  corrected SHAP scores for $C_2$.
\end{theorem}

\begin{proof}
  If $C_1,C_2$ are predictive equivalence then, by~\cref{prop:pe-xp},
  $C_1,C_2$ have equal sets of (W)AXps. Given the definition of
  characteristic function~\cref{eq:ccf}, then the corrected SHAP
  scores will be the same.
\end{proof}

As argued earlier in this section, in practice appears harder to
compute the Shapley values for \emph{every} instance than simply
deciding predictive equivalence.
\cref{prop:pe-sv} shows the connections that logic-based XAI reveals. 
Furthermore, \cref{prop:pe-sv} addresses one example of several
measures of importance that can be considered:

\begin{remark}
  Following the framework proposed in~\citep{lhams-corr24},
  \cref{prop:pe-sv} can be restated for several power indices
  studied in the context of a priori voting
  power~\citep{machover-bk98}, and generalized to the case of feature
  attribution in XAI by L\'{e}toff\'{e} et al.~\citep{lhams-corr24}.
\end{remark}

\subsection{Complexity Results Beyond DTs}

The computation of one abductive explanation is in P for
several families of classifiers, in addition to
DTs~\citep{iims-corr20,iims-jair22}.
This is the case with decision graphs, with or without canonical 
representations~\citep{hiims-kr21,hiicams-aaai22},
monotonic classifiers~\citep{msgcin-icml21},
naive bayes classifiers~\citep{msgcin-nips20},
restricted cases of multivariate decision trees~\citep{ccms-kr23},
and
$k$-nearest neighbors~\citep{barcelo-pods25}.
Regarding the computational problems studied in this paper, both
prediction with missing data and explanations are in P for the
families of classifiers listed above.

Regarding predictive equivalence, this paper (see~\cref{alg:pedt})
proves that the problem is in P for decision trees. The same algorithm
can be adapted for tractable fragments of multivariate decision
trees~\citep{ccms-kr23,cooper-pc25}.
While in general the comparison of representations for predictive
equivalence can be unsound if representations are non-canonical, this
is a correct approach for when classifiers exhibit canonicity. For
example, this is well-known to be the case with
$\{\bcf_c(C)\,|\,c\in\fml{K}\}$, i.e.\ the sets of prime implicants
for each predicted class for classifier
$C$~\citep{blake-phd37,brown-bk90,darwiche-jair02}. In 
addition, it is well-known that reduced ordered binary decision
diagrams (ROBDDs)~\citep{bryant-tcomp86,darwiche-jair02} and
sentential decision diagrams (SDDs)~\citep{darwiche-ijcai11} are
canonical
representations~\citep{bryant-tcomp86,darwiche-jair02,darwiche-aaai15}. As
a result, predictive equivalence can be decided in polynomial-time for
classifiers represented with ROBDDs and SDDs.

\section{Conclusions} \label{sec:conc}

Recent work by McTavish et al.~\citep{rudin-icml25,rudin-corr25}
demonstrated the importance of eliminating predictive equivalent
decision trees from the Rashomon set. Indeed, this work showed that a
significant percentage of decision trees in the Rashomon sets for
different classification problems are predictively equivalent.
Specifically, the work of McTavish et al.\ proposed an
approach~\cite{rudin-icml25,rudin-corr25} for solving a number of
computational problems of interest for working with Rashomon sets of
DTs, including deciding predictive equivalence of DTs.
At its core, the work of McTavish et al.\ consists of using the
well-known Quine-McCluskey (QM) method for computing minimimum-size
DNF representations of DTs. The additional algorithms proposed by
McTavish et al.\ run in linear time on the size of the computed
minimimum-size DNF.

In this paper, we show two critical issues, one related with the
efficiency of the algorithms  proposed by McTavish et
al.~\citep{rudin-icml25,rudin-corr25}, and the other with the use of
the QM method for predictive equivalence.
As a first issue, we show that the approach McTavish et al.\ is
unnecessarily complex in that it requires running the QM method for
solving a problem that is hard for the second level of the polynomial
hierarchy. In addition, we also prove that the worst-case exponential
running time and space of the QM method can be exercised in the case
of decision trees.
In contrast, in this paper, we propose novel polynomial-time
algorithms, in the size of the original DT, for \emph{all} of the
computational problems studied in the work of McTavish et
al.~\citep{rudin-icml25,rudin-corr25}.
The experiments confirm negligible running times in practice for all
the algorithms proposed in this paper.
As a second issue, we argue that the use of the set-covering step of
the QM method may produce incorrect results when deciding predictive
equivalence. In their work, McTavish et al.\ address the issue of
correctness by imposing constraints on how set-covering is
implemented. However, one of the constraints required by McTavish
et al.\ may be difficult to formally guarantee in practice; this will
depend on a number of aspects that our paper details.
%
%

Future work will investigate improvements to the asymptotic (but
polynomial) running times of the algorithms proposed in this paper,
but also assessing the alternatives revealed by the results
in~\cref{sec:xtra}. In addition, another topic of research would be to
devise asymptotically optimal algorithms for solving the computational
problems studied in~\cref{sec:ptime}.
Given the massive improvements that our algorithms yield, this line of
research is at present interesting mostly from a theoretical
perspective. 
Finally, \cref{sec:xtra} reveals several lines of research that are
also interesting from a theoretical standpoint. Future research will
address these lines of research.

\subsection*{Acknowledgments.}

McTavish et al.\ provided detailed comments to earlier versions of
this document.
This work was supported in part by the Spanish Government under
grant PID 2023-152814OB-I00, and by ICREA starting funds.
The first author (JMS) also acknowledges the extra incentive provided
by the ERC in not funding this research.

\vskip 0.2in

\newtoggle{mkbbl}

\settoggle{mkbbl}{false}

\iftoggle{mkbbl}{
    \bibliography{refs,dts}
}{

}

\end{document}